\documentclass[journal]{IEEEtran}
\ifCLASSINFOpdf
   \usepackage[pdftex]{graphicx}
   \graphicspath{{./images/}}
\else
\fi
\usepackage{amsbsy,amssymb,latexsym,amsmath}
\usepackage{cancel}
\usepackage{algorithm}
\usepackage{algpseudocode}
\usepackage{makecell}
\usepackage{subfigure}

\usepackage{caption}
\newcommand{\JOURNAL}{}

\begin{document}
%
\title{On Blocking Collisions between People, Objects and other Robots}
%

\author{Kwan~Suk~Kim,
        Luis~Sentis
\thanks{K.S. Kim is with the Department
of Mechanical Engineering at The University of Texas at Austin. E-mail: kskim@utexas.edu. L. Sentis is with the Department of Aerospace Engineering and Engineering Mechanics at The University of Texas at Austin. E-mail: lsentis@austin.utexas.edu}
\thanks{ This paper is supported by ONR N00014-15-1-2507 and NASA NRI NX12AM03G.}
}

\maketitle

\begin{abstract}
Intentional or unintentional contacts are bound to occur increasingly more often due to the deployment of autonomous systems in human environments. In this paper, we devise methods to computationally predict imminent collisions between objects, robots and people, and use an upper-body humanoid robot to block them if they are likely to happen. We employ statistical methods for effective collision prediction followed by sensor-based trajectory generation and real-time control to attempt to stop the likely collisions using the most favorable part of the blocking robot. We thoroughly investigate collisions in various types of experimental setups involving objects, robots, and people. 
Overall, the main contribution of this paper is to devise sensor-based prediction, trajectory generation and control processes for highly articulated robots to prevent collisions against people, and conduct numerous experiments to validate this approach. 
\end{abstract}


%
\IEEEpeerreviewmaketitle

\section{Introduction}
Is it okay for robots to stop objects or other robots that could collide with people? Although ground robots and autonomous cars have operated for a while in human populated environments, it is unusual to see them intervening over collisions that might happen externally to their current path. 

With today's advancements on autonomous systems, we are prompted to study such a problem, that of measuring the probability of an accident about to happen followed by a decision process for a robot to stop the likely event when physically possible. A posit here is that a robot might produce benefits to humans from three levels of safety: i) collision avoidance when possible, ii) in the case that collision avoidance fails, then collision detection and fast contact reaction, iii) in the case that a collision between an external object or robot and a person is likely to happen, then block the collision if physically possible. In this paper we explore case iii). And we have first studied this case from an engineering perspective without entering into ethical, moral, medical, psychological or social questions. These other questions are of course very important, but we wanted first to understand what methods could be devised and employed to intervene over external events using human-centered robots. For this study we use an upper body humanoid robot consisting of an articulated torso and two articulated arms. For sensing we use a structure light sensor mounted outside of the blocking robot. The point cloud sensor is able to simultaneously ``see'' the objects in the scene, the humanoid upper body robot, and people nearby. 

In order to make this study possible, the humanoid upper body robot needs two endow two capabilities: an estimation of the risk of a collision between nearby people and objects or robots, and an intervention strategy that is likely to block the object or robot if physically possible. We further give a twist to this study by considering that the blocking robot might be engaged into a task prior to considering intervening over a likely external collision. In that case, the robot must face the task of preventing the accident without stopping the task at hand. This question results in a study on constrained motion planning to engineer possible responses. Further details are broadly discussed next.
\begin{figure}%
\centering
\includegraphics[width=\linewidth]{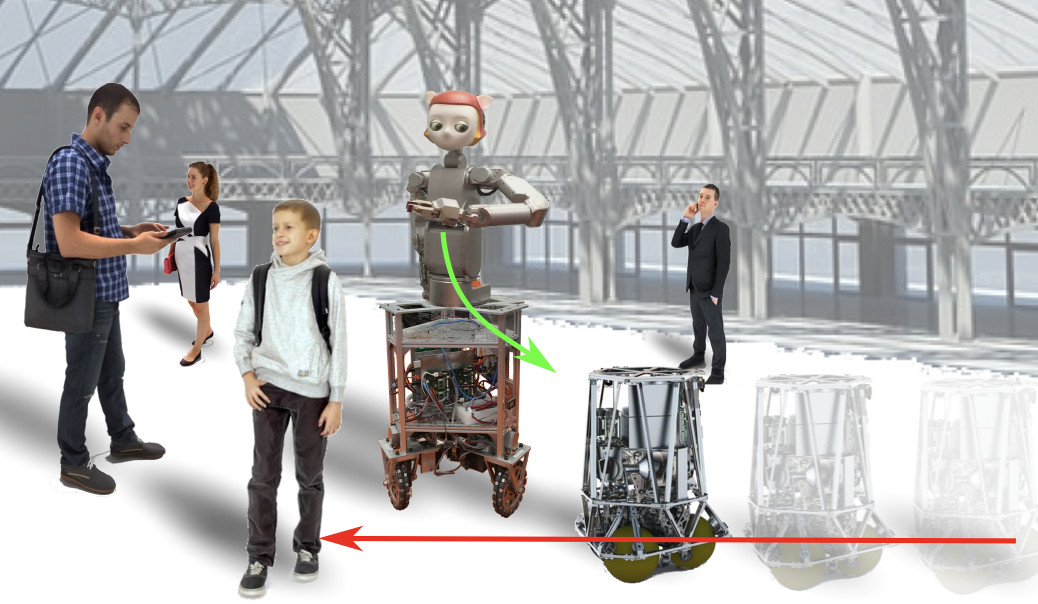}%
\caption
{{\bf Illustration of a collision intervention to prevent an accident.} 
	This scene depicts a mobile robot likely to hit a person unaware of the danger. A robot nearby is ready to intervene to stop the mobile robot for mitigating injury to the human.} \label{fig:intervention_scenes} 
\end{figure}

First, we use a point cloud sensor and prediction methods to estimate the risk that an object might collide with a person. Observing the current state of the surrounding objects to predict the future movements, the probability of collision of two objects or an object and a person is computed and the robot is controlled to prevent that collision from happening given a practical threshold risk value. Second, after it is decided that a collision prevention move should be attempted, it is necessary to compute intervention paths between any body part of the robot across its entire body and the estimated object path. For this purpose, we incorporate to our method a sampling-based motion planning algorithm and an analysis of the reachable volumes to determine the most favorable part of the robot's body and its path to intervene over the object's trajectory. We complement this computational processes with various experiments involving the humanoid upper body mentioned before blocking moving objects or robots using the humanoid robot's elbow, shoulder, torso, and end-effector in scenes where a human is likely to be struck by an object or by another robot. In light of this discussion, the main contribution of this paper is on devising sensor-based methods to effectively block collisions between external objects, robots and people using any part of a highly articulated robot, and while performing secondary tasks if physically possible. 


 \section{Background review}
Many studies have been devoted to robot safety, such as \cite{pervez2008} \cite{haddadin2009} \cite{vasic2013safety} \cite{haddadin2017}, focusing on mitigation methods between colliding robots and humans. However these studies do not consider blocking collisions between external objects or external robots and people.
The likeliness that a robot will collide with its environment has been widely studied to reason about the degree of safety and to create control policies for collision mitigation \cite{fraichard2004} \cite{weng2009toward} \cite{bouraine2011relaxing} \cite{du2011} \cite{levihn2014locally} \cite{cherubini2014autonomous} \cite{lesprance2014}.
Unlike our study the collisions anticipated in these studies are bounded to collisions between robots and their environments, without focusing on third-party collisions outside to a robot's path.

Robots approaching moving objects have been well studied by the robot grasping community \cite{lin1989} \cite{mehrandezh2000} \cite{aghili2012}. To grasp objects, a robot needs to estimate future object configurations. Some researchers predict future object configurations using a model of the object's dynamics \cite{mehrandezh2000} \cite{aghili2012}. But thanks to the recent innovations in machine learning, future object configurations can now be estimated without a prior model and used in a real-time process for grasping \cite{kim2012}. In contrast with grasping moving objects, our methods do not consider a particular appendage to interact with objects but instead perform a search over the entire articulated body of the robot.

As a twist on the research based on expected chores of human-centered robots, we assume that our blocking robot has a primary goal task prior to blocking moving objects. The control policy will attempt to find any other body part besides the task end-effector to block the moving object when physically possible. We will achieve this capability using sampling based motion planning methods \cite{kavraki1996} \cite{kuffner2011space}.
To fulfill a task priority, we will further employ the constrained motion planning method \cite{yao2005path} \cite{berenson2009} \cite{hauser2010multi} \cite{kim2016}. These methods have not been previously use to stop external objects or robots. Finally to perform these trajectories we will use whole-body operational space controllers on the intervening robot to enforce the task constraints via prioritized feedback linearization \cite{sentis2012}.

\section{Collision prediction}
\label{sec:colman}
\subsection{Object dynamics}
\label{sec:obj_dyn}
To simplify the prediction process, we consider only the positions of the observed objects, reducing them to simplified circular/spherical rigid bodies with no rotation. Then, we determine the configuration of each object via its center position. We assume that the linear acceleration of objects is produced by an unknown source with zero-mean normal distribution. The state of the i-th object at the m-th time slice (we assume a temporal discretization) is defined as follows.
\begin{align}
P_m^i \triangleq
\begin{pmatrix}
p_{o,m}^{i} \\ 
\dot{p}_{o,m}^{i} 
\end{pmatrix}
\end{align}
Also, each object is approximated with following dynamic equations including unknown disturbances $\xi_p^i$.
\begin{align}
P_m^i&= 
\underbrace{
\begin{pmatrix}
I & \Delta t I \\
0 & I
\end{pmatrix}
}_{A_p}
P_{m-1}^i
+
\xi_p^i \label{eq:st}
\end{align}
We assume that the position of the i-th object at time index $m$, $y_m^i$, has noise, $\xi_s$, with a normal distribution as follows.
\begin{align}
y_m^i &= 
\underbrace{\begin{bmatrix}
I & 0 \end{bmatrix}}_{C_p}
P_m^i + \xi_s
\end{align}
Then, the state of the i-th object and its covariance matrix can be estimated via a Kalman filter as follows.
\begin{align}
\tilde{P}_m^i &= \hat{P}_{m}^i + K_m^i \left( y_m^i - C_p \hat{P}_{m}^i\right) \\
\tilde{\Phi}_m^i &= \left( I - K_m^i C_p \right) \hat{\Phi}_m^i
\end{align}
where all new variables are defined in Table \ref{tab:kalman}.

We define a new random variable to estimate the future state of objects, $X_k^i$, and its initial value can be determined from the result of the Kalman filter as follows. 
\begin{align}
X_0^i & \sim \mathcal{N} \bigg( 
\tilde{P}_m^i
,~ 
\Phi_m^i
\bigg) 
\end{align}
Given the initial state and its state transition equation, Eq. (\ref{eq:st}), the random variable $X_k^i$ can be expressed using the following normal distribution.
\begin{align}
\begin{split}
X_k^i &= \begin{pmatrix} x_k^i \\ \dot{x}_k^i \end{pmatrix}  
\sim
\mathcal{N} \Bigg( 
\underbrace{
\begin{pmatrix}
\overline{x}_{k}^i \\
\dot{\overline{x}}_{k}^i
\end{pmatrix}
}_{\overline{X}_k^i}
,~
\underbrace{
\begin{pmatrix}
\Sigma_{x,k} & \Sigma_{xv,k}^i \\ \Sigma_{vx,k}^i &  \Sigma_{v,k}^i 
\end{pmatrix}
}_{
\Sigma_{X,k}^i
}
\Bigg) 
\\
&=
\mathcal{N} \left( 
A_p \overline{X}_{k-1}^i
,~
A_p 
\Sigma_{X,k-1}^i
A_p^{T}
+
\Sigma_w^i 
\right) 
\label{eq:X}
\end{split}
\end{align}
where $\overline{X}_k^i$ is the expected object state. In the state prediction, the time index $k$ at the current time is set to zero. In Appendix \ref{sec:simple}, the random variables are approximated with normal distributions for computational efficiency.

We now define a conditional random variable, $X_f^{ij} \triangleq (x_f^{ij},\, \dot{x}_f^{ij})$, representing the state of the i-th object under the condition that it has not collided yet with the j-th object. The probability distribution of the random variable is the complement of the collision probability between the i-th and j-th objects, so $X_f$ is not a normal distribution. $X_f$ also propagates in time similarly to $X$ in Eq. (\ref{eq:X}) such that we can define another random variable, $\hat{X}_f$ which is a one-step propagation process given $X_f$. Even though $X_f$ and $\hat{X}_f$ are not normal distributions, the probabilistic properties of $\hat{X_f}$ can be derived from those of $X_f$ as shown in Appendix \ref{sec:xf}.
\begin{align}
\begin{split}
{\rm E} \left( \hat{X}_{f,k} \right) &= A_p {\rm E} \left( X_{f,k-1} \right) \\
{\rm V} \left( \hat{X}_{f,k} \right) &= A_p {\rm V} \left( X_{f,k-1} \right) A_p^T + \Sigma_w \label{eq:Xhat}
\end{split}
\end{align}
%

\subsection{Instantaneous collision probability} 
\label{sec:ic}
Based on the object dynamics of Eq. (\ref{eq:st}), we can predict the future locations of moving objects and also determine whether a collision external to a blocking robot will take place. 
By observing the current state of objects and their covariances, we can anticipate their future state, and estimate the possibility of their collision as stochastic processes as shown in Eq. (\ref{eq:X}).

For convenience, all the probability density functions and probabilities are denoted as $\mathbf P$ and $\mathbf p$, respectively, and all the predicates are denoted as $\mathbf q$. We define a probability function, $\mathbf{P}_o^i: \mathbb{R}^{\rm N_d} \rightarrow \mathbb{R} $ being the probability that the center of the i-th object is located at a given point, $p^i$, as follows.
\begin{align}
\mathbf{P}_o^i\left(p^i\right) \triangleq \mathbf{P}( x^i = p^i )
\end{align}
where $x^i$ is the center position of the i-th object. Also, a predicate prescribing whether the i-th and j-th objects collide with each other is labeled as $\mathbf{q}_{ic}(i,j)$.

Before computing the cumulative collision probability over time, we need to consider the probability of collision at a given time and the probability density function associated with object positions. The probability that the i-th object located at $p^i$ collides with the j-th object can be derived from the probability density function of the i-th object, $\mathbf{P}_o^i$ and that of the j-th object, $\mathbf{P}_o^j$, and it can be expressed as $P_{ic}$ as follows. 
\begin{align}
\begin{split}
&{P}_{ic} ^{ij}\left( p^i, \mathbf{P}_o^i, \mathbf{P}_o^j \right) \\
\triangleq& 
\mathbf{P}_o^i\left(p^i\right)\, 
\mathbf{P}\left( f_c^{ij}(p^i, p^j) = 1 \Big| p^j \sim \mathbf{P}_o^j \right) \\
=& 
\mathbf{P}_o^i\left(p^i\right)\, 
\int_{\mathcal{S}} f_c^{ij}(p^i, p^j) 
	\mathbf{P}_o^j\left(p^j\right) dp^j \label{eq:pic} 
\end{split}
\end{align}
where 
$f_c^{ij}$ is a function that checks collision events between the i-th and j-th objects, mapping the positions of the two objects to value $1$ if the objects collide with each other or $0$ otherwise.
The probability density function, $\mathcal{S}$, corresponds to the probability that two objects collide with each other. The collision probability can be determined by the density functions at a given time instance, so we call it the {\it instantaneous collision probability}, $\mathbf{p}_{ic}^{ij}$. 
\begin{align}
\mathbf{p}_{ic}^{ij} \triangleq \int_{\mathcal{S}} {P}_{ic} ^{ij}\left( p^i, \mathbf{P}_o^i, \mathbf{P}_o^j \right) dp^i \label{eq:pic2}
\end{align}
	
We have derived the probability that two objects collide with each other. Therefore, we can also derive the probability that they do not collide with each other, and the conditional density function of the collision-free object positions. The collision density function of Eq. (\ref{eq:pic}) is the probability that the i-th object located at $p^i$ collides with the j-th object. The corresponding probabilistic density function describing that the i-th object at $p^i$ is free from colliding with the j-th object is represented as $\mathbf{P}_{of}^{ij}$ as follows.
\begin{align}
\mathbf{P}_{of}^{ij} \left( p^i, \mathbf{P}_o^i,\, \mathbf{P}_o^j \right)
\triangleq
\frac{
\mathbf{P}_o^i \left( p^i \right) - {P}_{ic}^{ij} \left( p^i,\, \mathbf{P}_o^i,\, \mathbf{P}_o^j \right)\label{eq:pof}
}
{ 1 - \mathbf{p}_{ic}^{ij} }
\end{align}
where $\mathbf{P}_o$ is the probability density function of the collision-free object at a given time. Therefore, Eq. (\ref{eq:pof}) can be indexed at time $k$ as follows.
\begin{align}
\mathbf{P}_{of,k}^{ij} \left( p^i \right)
\triangleq
\frac{
\hat{\mathbf{P}}_{of,k}^{ij} \left( p^i \right) - {P}_{ic}^{ij} \left( p^i,\, \hat{\mathbf{P}}_{of,k}^{ij},\, \hat{\mathbf{P}}_{of,k}^{j} \right)\label{eq:pofk}
}
{ 1 - \mathbf{p}_{ic}^{ij} }
\end{align}
where $\hat{\mathbf{P}}_{of,k}$ is the collision-free probability density function at time $k$ which is predicted from that at time $k-1$. From the definition of $X_f$ and $\hat{X}_f$ in Sec. \ref{sec:obj_dyn}, $\mathbf{P}_{of}$ and $\hat{\mathbf{P}}_{of}$ correspond to their probabilistic density functions.  
\begin{align}
\begin{split}
X_f &\sim \mathbf{P}_{of} \\
\hat{X}_f &\sim \hat{\mathbf{P}}_{of}
\end{split}
\end{align}
If $\mathbf{P}_{of,k-1}$ is a normal distribution, $\hat{\mathbf{P}}_{of,k}$ can be estimated from Eq. (\ref{eq:X}). Though $\mathbf{P}_{of}$ is not a normal distribution, we approximate it to be a normal distribution in Appendix \ref{sec:simple}.
\begin{figure}[t]\centering
\ifdefined\JOURNAL
\includegraphics[width=\linewidth, clip=false ] {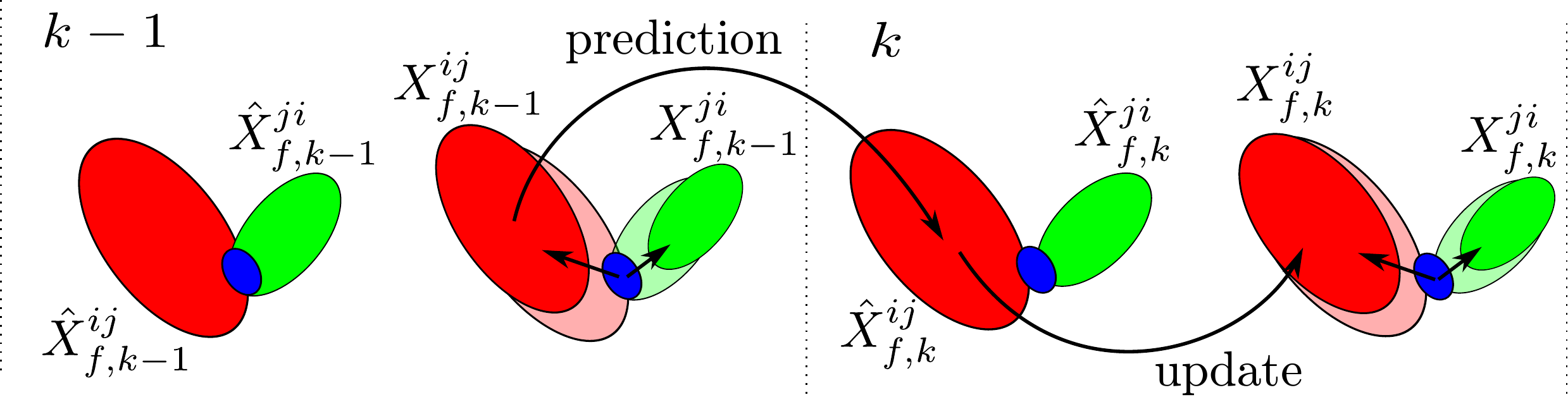}
\else
\includegraphics[width=1.0\linewidth, clip=false ] {x_free.pdf}
\fi
\caption
[ Prediction and update of the collision-free random variable, $X_f$]
{{\bf Prediction and update of the collision-free random variable, $X_f$.}
During the prediction process, the distribution of the object positions diverge to $\hat{X}_f$ because of the uncertainties given by Eq. (\ref{eq:X}), and the update using the complement of the instantaneous collision probability. 
} \label{fig:x_free} 
\end{figure}



\subsection{Cumulative collision probability}
The collision probability given in Eq. (\ref{eq:pic2}) estimates only the instantaneous probability at a given time, so we cannot use that probability to determine how likely a collision happens at that time because a collision might already have happened and the objects involved in it could keep colliding. Thus, we need to recursively accumulate the collision probability and exploit the conditional probability that collisions do not happen before a time horizon. The predicate that describes this situation can be defined as follows.
\begin{align}
\mathbf{q}_{ac,k}(i,j) &\triangleq \bigvee_{n \le k} 
\left\| x_{n}^i - x_{n}^j \right\| \le r^i + r^j 
\end{align}
The corresponding accumulated collision probability is expressed and computed as follows.
\begin{align}
\mathbf{p}_{ac,k}^{ij} 
=& ~\mathbf{p} \left( \mathbf{q}_{ac,k} (i,j) \right)  =~\mathbf{p}_{ac,k-1}^{ij} + 
\label{eq:pac}
\\
&	~	\big( 1 - \mathbf{P}_{ac,k-1} ^{ij}\big) 
\mathbf{p}\left(  
		\mathbf{q}_{ic,k}\left(i,j\right) \bigg| \neg \mathbf{q}_{ac,k-1}\left( i,j\right) \right) 
\end{align}
The conditional probability on the last term of Eq. (\ref{eq:pac}) is equivalent to the integral of the collision probability of Eq. (\ref{eq:pic}), yielding the following.
\begin{align}
\mathbf{p}_{ac,k}^{ij} 
=&		~\mathbf{p}_{ac,k-1}^{ij} + \nonumber\\
&	~	\big( 1 - \mathbf{p}_{ac,k-1} ^{ij}\big) 
\int_{\mathcal{S}}
{P}_{ic}^{ij} \left( p^i, \mathbf{P}_{of}^{ij}, \mathbf{P}_{of}^{ji} \right) dp^i
\label{eq:pac2}
\end{align}

The cumulative collision probability, $\mathbf{p}_{ac}$, increases over time and can be computed recursively starting at time zero. However, the derivation of the cumulative collision probability, $P_{ic}$, based on non-normal distributions is computationally expensive, prompting us to use an approximation.
Thus, the probability is simplified for real-time computation purposes in Appendix \ref{sec:a_pc}.
Given the cumulative collision probability and a probability threshold, $\eta$, we can derive the minimum time index, $k_c$, at which the probability of a collision between any two objects from taking place exceeds the threshold as follows.
\begin{align}
&k_c = \text{argmin} \; k \nonumber \\
&\text{such that} \nonumber\\
&\bigvee_{i,\,j \le \rm N_o} 
\bigg( \left( i \neq j \right) \wedge \Big(\mathbf{p}_{ac,k}^{ij} \ge \eta \Big) \bigg) \label{eq:tc2}
\end{align}
$k_c$ embodies the likelihood that at least one object pair in the environment will collide with each other at time $k_c\cdot \Delta t$ with a probability $\eta$.




\begin{figure}[t]\centering
\ifdefined\JOURNAL
\includegraphics[width=\linewidth, clip=false ] {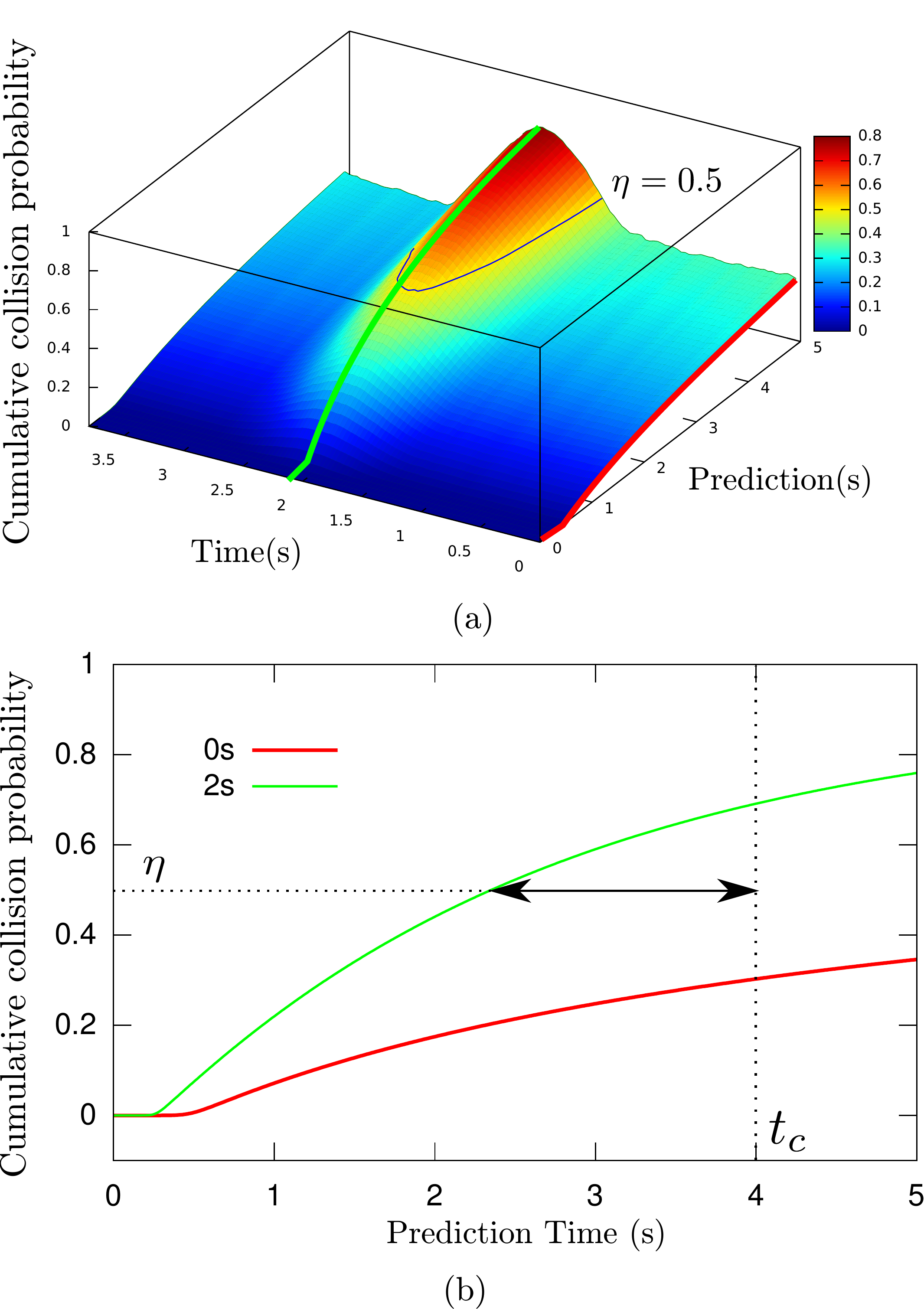}
\else
\includegraphics[width=0.7\linewidth, clip=false ] {cum_col2.pdf}
\fi
\caption
[ Cumulative collision probabilities over time] 
{{\bf Cumulative collision probabilities over time}: this image 
	shows the estimated cumulative collision probabilities when an object is getting closer to another one (0-2sec) and receding from it (2-4sec). Whenever a new observation is produced by an external sensor, the agent computes the cumulative probabilities. Fig. (b) shows the collision predictions at time 0 and 2, which are the slices in Fig. (a).
} \label{fig:cum_col} 
\end{figure}

\section{Behavior planning}
\label{sec:agent}

\begin{figure}\centering
\ifdefined\JOURNAL
\includegraphics[width=0.9\linewidth, clip=false ] {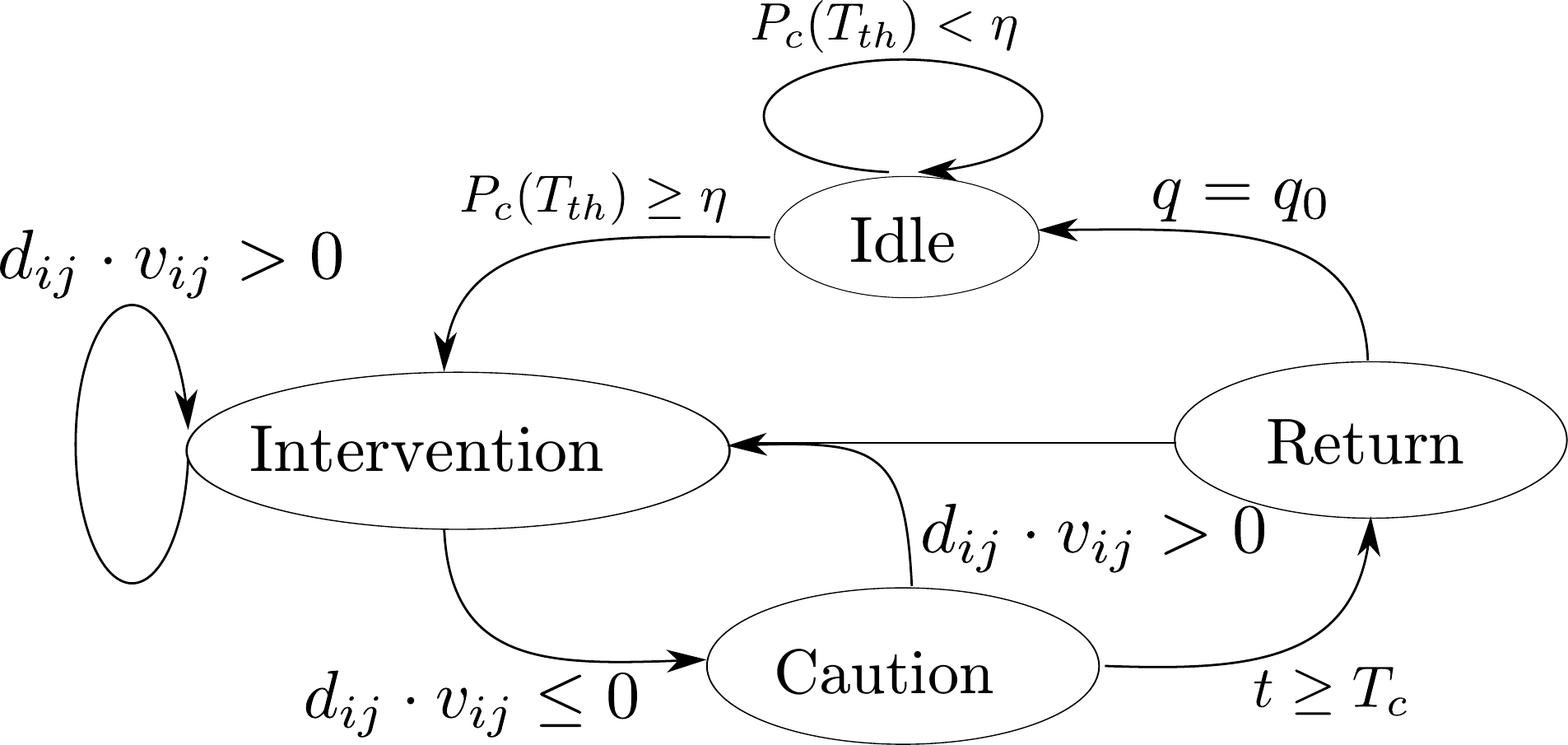}
\else
\includegraphics[width=0.7\linewidth, clip=false ] {col_fsm.pdf}
\fi
\caption
[ State transition diagram]
{{\bf State transition diagram}. Once the collision probability exceeds a given threshold, the blocking robot triggers an intervention mode.
} \label{fig:col_fsm} 
\end{figure}

To respond to the anticipated collisions, the robot agent uses four operating modes: idle, intervention, caution, and return. In the idle mode, the agent keeps predicting collisions and continuously planning a potential motion to block the most likely collision, but does not start an intervention maneuver in this mode. In the intervention mode, the agent uses the most favorable body part to intercept the likely colliding external objects with the hope to deviate one of them from the set trajectory. In the caution mode, the agent expects that anticipated collisions are unlikely but have enough risk. It then creates approaching behaviors to the threatening objects with its body parts without touching them. Finally, in the return mode, the agent assumes that there are no collision threats and returns to its normal operation. The modes and their transitions are described in the state diagram of Fig. \ref{fig:col_fsm}


\subsection{Idle mode}
The cumulative probability of a collision at every instant over a prediction horizon asymptotically increases as suggested by Eq. (\ref{eq:tc2}). However, the time when the probability exceeds a given threshold depends on the initial object states and the statistics of the random variables. If a collision will happen in the distant future, the agent does not need to be concerned. Otherwise, the robot agent, according to premises, will intervene and attempt to stop the collision between the two external objects. Therefore, a probability threshold and a time threshold to differentiate the distant and the near futures need to be considered to reason about interventions. In Fig. \ref{fig:cum_col}, the probability threshold, $\eta$ and the time threshold, $T_{th}$ are characterized. If the collision probability of any two object at $T_{th}$ exceeds the probability threshold, $\eta$, then the agent intervenes by switching its mode from idle to intervention mode.
Therefore, if any pair of objects are expected to collide with each other within the set time period, the mode of the robot agent changes as follows.
\begin{align}
\Bigg( \bigvee_{i,\,j \le \rm N_o,\ i \neq j} 
\Big(P_{ac,T_{th}/\Delta t}^{ij} \ge \eta \Big) \bigg) \label{eq:trans1}
\nonumber \\ 
\rightarrow
\left( 
mode \gets \rm INTERVENTION	
\right)
\end{align}
The pair of objects for which the collision probability exceeds the threshold at $T_{th}$ forms a new set, $\mathcal{P}_c$, corresponding to the set of imminent collisions as follows. 
\begin{align}
\mathcal{P}_c = \left\{  \left(i,j\right) \bigg| P_{ac,T_{th}/\Delta t}^{ij} \ge \eta \right\} \label{eq:pc_set}
\end{align}

\subsection{Intervention mode}
In the intervention mode, the agent attempts to intercept the trajectory of the most imminent collision event. The most imminent collision is chosen from the set $\mathcal{P}_c$ defined in Eq. (\ref{eq:pc_set}) by considering the positions and velocities of the external objects. Each potential collision consists of two objects, and the time when these two objects will make contact or come the closest to each other can be derived from their initial state. Let the position and velocity of the i-th object be $p_i$ and $v_i$, respectively, then the time when the distance between them is minimum can be expressed as follows.
\begin{align}
\frac{d}{dt} \Big\| \left(p_i + v_i\, t\right) - \left(p_j + v_j\, t \right) \Big\| = 0 \\
\left( p_i - p_j \right) \cdot \left( v_i - v_j \right) < 0
\end{align}
The time at which the i-th and j-th object approach the closest or collide with each other is defined as $t_c^{ij}$ and can be derived analytically as follows.
\begin{align}
t_c^{ij} =  
- \frac{\left( p_i - p_j \right) \cdot \left( p_i - p_j \right)} 
{\left( p_i - p_j \right) \cdot \left( v_i - v_j \right)} 
\label{eq:xxx}
\end{align}
Then the pair of objects that will be involved in the most imminent collision can be derived as follows.
\begin{align}
( i_c, j_c ) = \underset{
	\left( i,j\right)}
	{\rm argmin} \ 
	t_c^{ij}, \left(i,j\right) \in \mathcal{P}_c 
\end{align}
Once the pair of objects involve in the most imminent collision is identified, the positions and velocities of the objects are used to determine the robot agent's motion to intercept them. From that pair, the time when they come the closest to each other is derived via Eq. (\ref{eq:xxx}).
Subsequently, a motion planner described in the next section generates a path for the robot agent to intercept the line segments of the external moving objects along the colliding trajectory to be executed. Since the most imminent collision is identified with our method and the robot agent's mode switches to the intervention mode, the agent keeps updating the expected trajectories of the external objects and determines whether to remain in the intervention mode. The decision to stay in that mode considers the collision probability and the inner product of the relative position and velocity of the two objects. If the probability is still higher than the threshold or the inner product is negative \-- meaning that the distance between the objects decreases \-- then the agent remains in the intervention mode. Otherwise, the mode switches to caution.
\begin{align}
\Bigg( \bigwedge_{i,\,j \le \rm N_o, i \neq j } 
\Big(P_{ac,T_{th}/\Delta t}^{ij} < \eta \Big)  
\wedge \Big( d_{ij} \cdot v_{ij} \le 0 \Big)
\Bigg) \nonumber \\ 
\rightarrow
\left( 
mode \gets \rm CAUTION	
\right)
\label{eq:trans2}
\end{align}

Once the mode switches to intervention status and the most imminent collision is determined, the external object pair expected to be intercepted by the robot agent will not change until the estimated collision probability becomes unlikely due to the blocking effect of the interception. The agent will fully complete the intervention for a particular collision even though a more imminent collision is newly identified.

\subsection{Caution mode}
If the collision probability between the two external objects is lower than the set threshold and the two objects are determined to move away from each other, we assume that a collision is unlikely. If this happens during the idle mode, the agent will remain on that state. However, if it happens during the intervention mode, the robot agent will change its policy to caution mode because the intervention process in which it is involved in will be less urgent.

Instead of "forgetting" the intervention task the agent started, it keeps being "concerned" about it by using the caution mode. In the caution mode, the agent neither intervenes nor returns to fully focusing on the primary goal task. Instead it stays ready for intervention and if all of a sudden the collision probability increases, the robot agent goes back to the intervention task.
\subsection{Return mode}
When the likelihood of a collision is low within the set time horizon, $T_{ca}$, based on Eq. (\ref{eq:trans2}) the agent goes back to its normal operation consisting on executing some primary task other than blocking collisions. To return to a set default posture for normal operation, the robot switches its logic mode to return. 
The transition condition to trigger the return mode is shown below.
\begin{align}
\left( 
mode = \rm CAUTION	
\right) \wedge
\Bigg( \bigwedge_{i,\,j \le \rm N_o, i \neq j } 
\Big(P_{ac,T_{th}/\Delta t}^{ij} < \eta \Big)  
\Bigg) \nonumber \\ 
\wedge ( t > T_C) 
\rightarrow
\left( 
mode \gets \rm RETURN	
\right)
\label{eq:trans3}
\end{align}

\section{Motion planning}
As stated, the robot agent is set to intervene over likely external collisions by using the most favorable body part available over its entire body, and based on the behavior triggers described in Sec. \ref{sec:agent}. Although there are many types of robot "anatomies" and types of appendages, in this study we consider an articulated upper body torso with an anthropomorphic. The structure has $N_j$ joints and $N_l$ links and is kinematically redundant for performing end-effector tasks in SE(3) or $\mathbb{R}^3$. In our formulation, the robot agent will attempt to use its most favorable body part to block collisions between pairs of external objects, and it will do so by exploiting its redundant kinematic abilities. To fulfill a primary task supposedly being performed with the end-effector, any of the robot agent's body parts other than the end-effector will be explored to block external collision unless there is no other choice other than giving up the primary task and employing the end-effector. To achieve such behavior, we leverage the method of reachable volumes where we simultaneously consider the reachable space for all body parts, and search in the robot's configuration space for postures that fulfill the primary and the blocking tasks. 

Subsequently, a we employ a motion planner to generate configuration space trajectories that achieve the desired goals. Since we are dealing with multiple objectives, the reachable volumes and the trajectories are derived using constrained planners. Even though the agent has redundant degrees of freedom to exploit the null space of the end-effector task, reaching the collision target may still turn to be physically impossible. In that case, the robot agent will be prompted to give up on the end-effector task and attempt to block the collision without task constraints. Because of this dichotomy, we will both compute constrained and unconstrained reachable motion plans.

Due to the need for real-time interventional speed, both trajectory planning and execution should be achieved within a set time horizon that increases the chances for success. To achieve this, we split the motion planning process into an off-line and an on-line procedures similar to the two-phase motion planning process discussed in \cite{kavraki1996}. During the off-line process, reachable volumes of all joints and linkages of the blocking robot are generated, and the configurations that correspond to the volumes are interconnected as tree structures according to the probabilistic roadmap (PRM) method \-- see \cite{kavraki1996}. During roadmap generation, dynamic properties associated with the configurations are also computed.

\subsection{Computing reachable volumes}
\begin{figure}\centering
\ifdefined\JOURNAL
\includegraphics[width=0.8\linewidth, clip=false ] {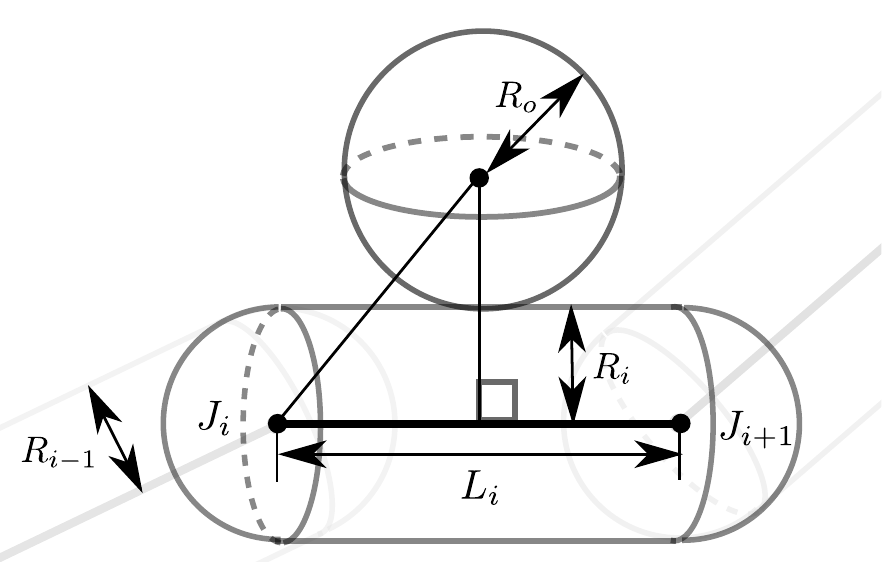}
\else
\includegraphics[width=0.6\linewidth, clip=false ] {int_link.pdf}
\fi
\caption
[ A configuration in which a $i$-th link makes a contact with an external object of radius $R_o$] 
{{\bf Shown here is a robot's configuration in which an $i$-th link makes a contact with an external object of radius $R_o$} 
} \label{fig:int_l} 
\end{figure}

To use any of the blocking robot's body parts, their position needs to be evaluated during the motion planning process. We approximate each part as two hemispheres connected with a cylinder as shown in Fig. \ref{fig:int_l}.

\begin{figure}\centering
\includegraphics[width=\linewidth, clip=false ] {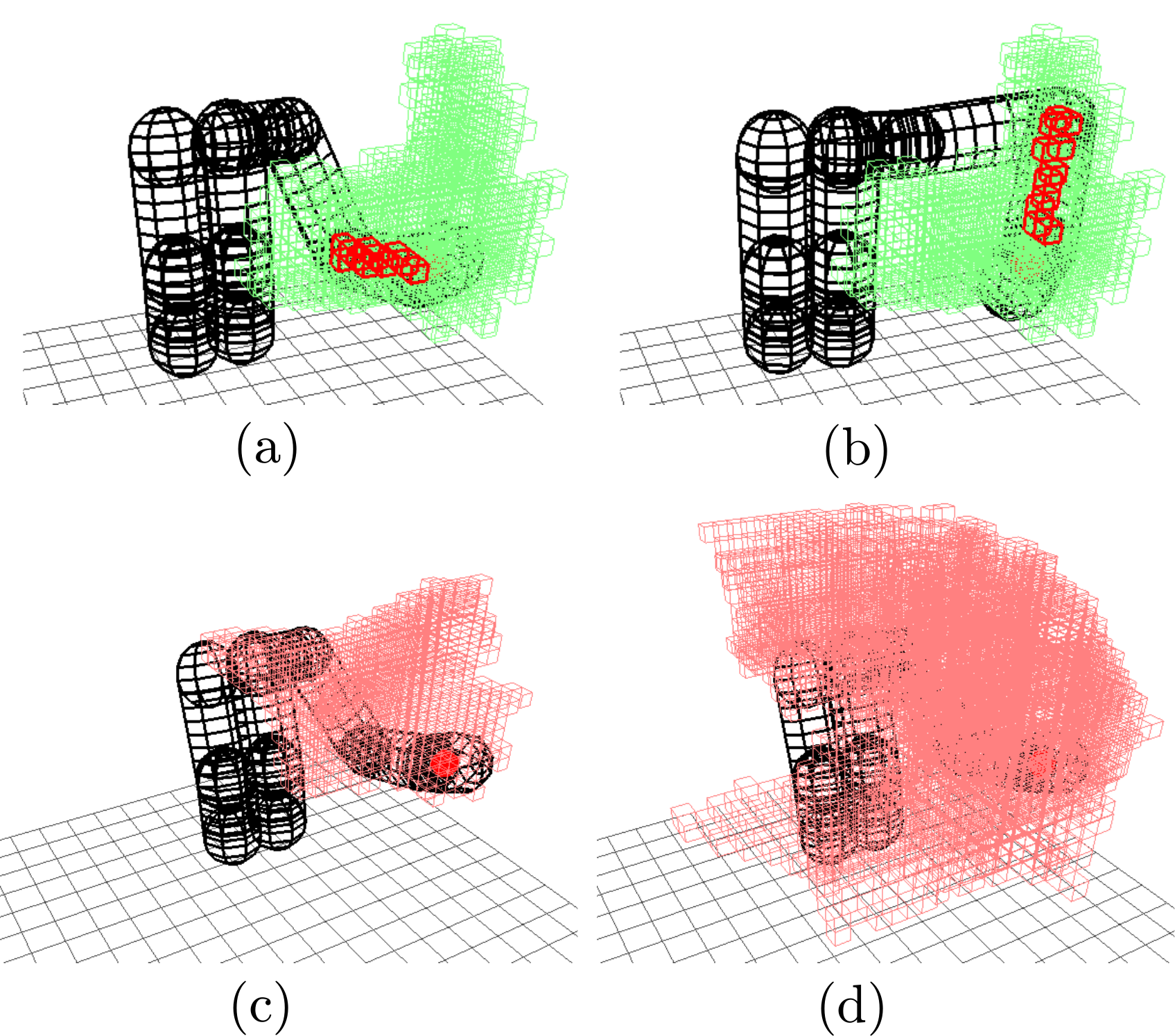}
\caption
{{\bf The reachable volume}
corresponding to a robot's upper arm while fulfilling an end-effector task is generated via an occupancy map of all possible postures as shown in (a) and (b). The constrained and unconstrained reachable volumes of the robot's upper arm compared in (c) and (d). The location of the end-effector for computing constrained reachable volumes is assumed to be stationary. The unconstrained case does not impose end-effector task location requirements. The red ball shown at the end-effector depicts its goal position used for the constrained scenario.}
 \label{fig:reavol} 
\end{figure}

To enable real-time interventional performance while exploring the aforementioned kinematic combinatorics, we generate reachable volumes for each body linkage a priori as described in \cite{mcmahon2014}. In contrast, our reachable volumes are generated by sampling feasible configurations rather than analytically computing the Minkowski sum of the reachable volumes. This approach allows us to deal with more complex shapes and joint limits. We first model each link as a line segment and generate reachable volumes as shown in Fig. \ref{fig:reavol}.
Adjacent Cartesian coordinates of the link positions are clustered, and a 3D cell array structure is generated similarly to \cite{csucan2012}. Each cell represents a set of adjacent points in 3D space and consists of the configurations for which the body link positions are located in the cell. We use an octree \cite{meagher1980} to efficiently represent the reachable volumes. There are $N_l$ octrees for the constrained space and another $N_l$ octrees for the unconstrained case. The computation time for building octrees is $\mathcal{O} ( \log N )$, which is considered as moderately expensive. The algorithm for building octrees is shown in Algorithm 1.
\begin{algorithm}[t]
\caption*{Algorithm 1. Building octree}
\begin{algorithmic}
\Function {\rm BuildOctree}{ $Link$, $Octree$ }
\State $p_1 \gets Link.from$
\State $p_2 \gets Link.to$
\State $r \gets Link.radius$
\If { $Octree.depth$ = \rm MAX\_DEPTH }
\State {$Octree.links \gets Link$}
\State \Return
\EndIf
\If { $Octree.depth = \emptyset$ }
\State {\rm CreateChildren($Octree$) }
\EndIf
\For {$child$ in $Octree.children$}
\State { $d \gets $ {\rm getDistToLineSeg}($child.center,\, p_1,\,p_2)$}
\If { $d \le  child.width + r$ }
\State { \rm BuildOctree} ( $Link$, $child$ )
\EndIf
\EndFor
\State \Return
\EndFunction
\end{algorithmic}
\end{algorithm}

The sampled configuration fed into the Octree algorithm is first generated by the PRM method {\cite{kavraki1996}. As mentioned, constrained configurations are those which keep the end-effector at a desired task location while unconstrained configurations do not impose any requirement. To consider both constrained and unconstrained options, we generate two different roadmaps. The algorithm for roadmap generation is shown in Algorithm 2. In there, the variable $Link$ consists of four tuples, $from$, $to$, $r$, $q$ corresponding to the position of one end of the link, the position of the other end, the radius of the link, and the configuration to which the link belongs to, respectively. Algorithm 2 is similar to Algorithm 1, but in the former, the sampled configuration is used to generate the reachable volumes. Also, in the case of constrained motion planning, we exploit the null-space projection of the primary goal task to fulfill the primary end-effector task. The detailed explanation of the constrained planner is described in Sec. \ref{sec:wbmp}. The constrained and unconstrained reachable volumes of our robot's left upper arm are shown in Fig. \ref{fig:reavol}-(c) and (d).
Whenever a new sampled configuration is considered, the sample is registered as a new node in a search tree. In addition, the Cartesian coordinates of each link of the sampled configuration is also registered in a corresponding 3-dimensional cell of the reachable volume. The roadmap also provides paths to connect all configurations, so the motion plan from the current state to the desired interventional state can be effectively derived from the roadmap.
\begin{algorithm}[t]
\caption*{Algorithm 2. Learning phase}
\begin{algorithmic}
\Function {\rm Learn}{}
\State $N$ $\gets$ $\emptyset$,  
$E$ $\gets$ $\emptyset$,
$O$ $\gets$ $\emptyset$
\While{ 1 }
\State $q_{new}$ $\gets$ RandomConf()
\State $q_{near}$ $\gets$ NearestNeighbor($q_{new}$, $N$)
\State $dq$ $\gets$ Project($q_{near}$, $q_{new} - q_{near}$) 
\State $\Delta q$ $\gets$ $dq / \left|dq \right| \times \bf STEP$
\State $q_{prev}$ $\gets$ $q_{near}$
\State $q$ $\gets$ $q_{prev}$ + $\Delta q$
\While{ IsFeasible($q$) }
\State $N$ $\gets$ $N \cup q$
\State $E$ $\gets$ $E \cup \left( q,\, q_{prev}\right)$
\For {$l_i$ in {Links($q$)} }
\State {BuildOctree($l_i$, $O_i$)}
\EndFor 
\State $q_{prev}$ $\gets$ $q$
\State $q$ $\gets$ $q$ + $\Delta q$
\EndWhile
\EndWhile
\EndFunction
\end{algorithmic}
\end{algorithm}


\subsection{Searching for intervening body parts}
During the collision intervention process described in Sec. \ref{sec:agent}, the blocking robot needs to determine which body part can intercept the predicted external collision. Because the reachable volumes of all the linkages of the blocking robot are computed in advance, the feasible configurations for intervention can be determined by searching the reachable volumes. Given an object position, $p_o$, its velocity, $v_o$, and its radius, $r_o$, we can define the set of configurations, $C_{int,j}$ at which the j-th link intersects the trajectory as follows.
\begin{align}
\begin{split}
C_{int,j} = \big\{& l_k.q \in C \ \bigg|\ l_k \in O_j.links,\ \\ 
&Distance(l_k.from,\, l_k.to,\, p_o,\, v_o \cdot t_d ) \le r_o + r \big\} \label{eq:c_intj}
\end{split}
\end{align}
where the function, $Distance$ returns the minimum distance between two line segments. The four arguments for that function are the pairs defining the line segments. The configurations of the previous set have at least one link that intersects the object trajectories. If the set has at least one element, the robot can intervene to stop the collision using the corresponding body link. Our robot consists of $N_l$ body linkages, and we consider constrained and unconstrained motion plans. Therefore, given a collision trajectory, $2 N_l$ sets of intersecting configurations need to be considered.

The decision policy for the intervening body linkage is shown in Fig. \ref{fig:decision}.
\begin{figure*}\centering
\includegraphics[width=\linewidth, clip=false ] {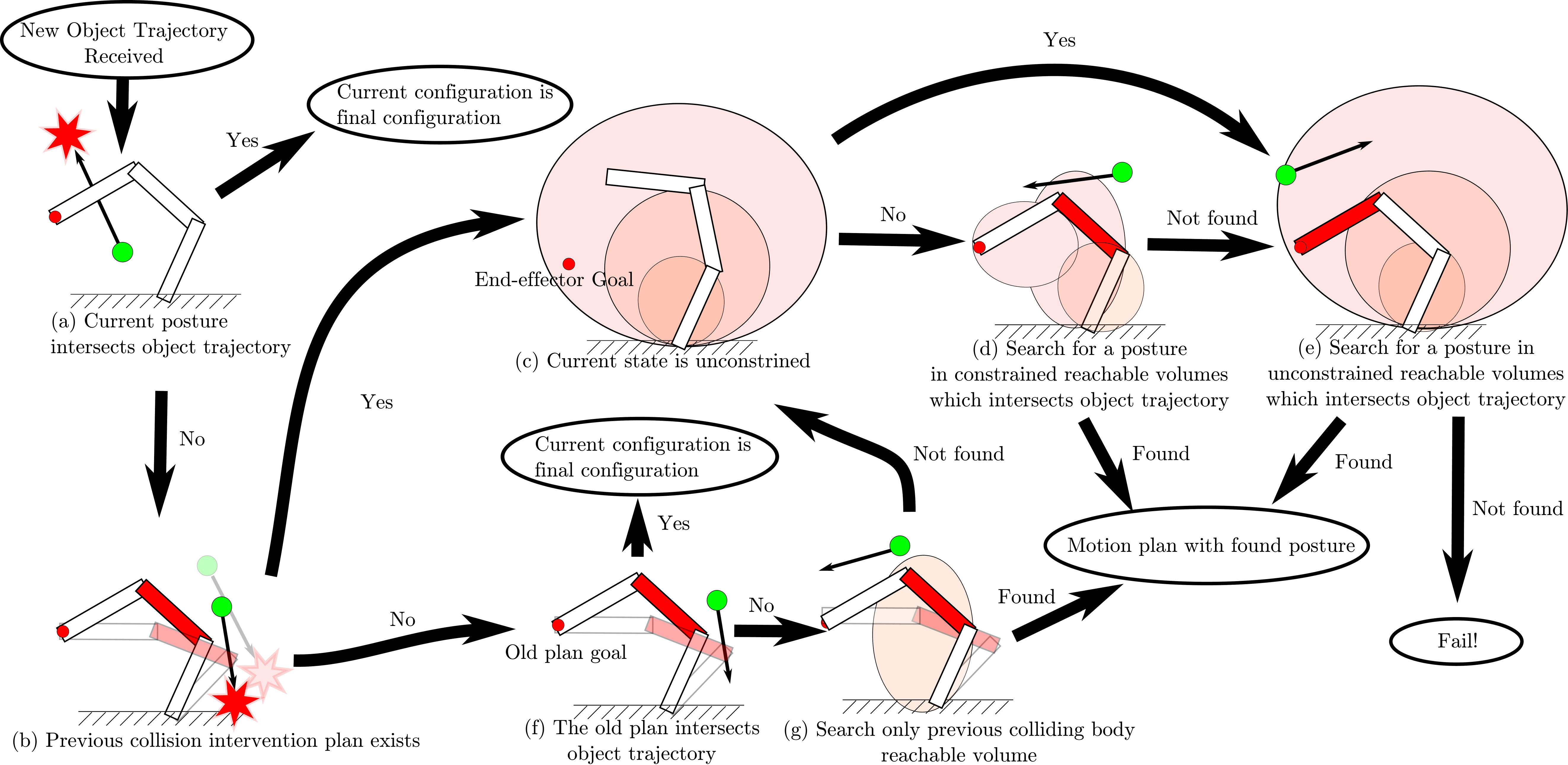}
\caption
[Decision policy for the motion planner process]
{{\bf Decision policy of the intervening body linkage}. Given a computed motion plan, the blocking robot will reuse it. Otherwise, a new motion plan is generated.
} \label{fig:decision} 
\end{figure*}
Whenever the state of the external objects changes, the blocking robot determines new intervening body parts. If the current robot configuration intersects the trajectory of the object (Fig. \ref{fig:decision} (a)), then it should remains at that configuration. 
Otherwise, the decision process checks whether the blocking robot is operating in the intervention mode (Fig. \ref{fig:decision} (b)) such that it can decide whether to reuse the previous motion plan. If the robot is already in intervention mode, the final destination given by the previous motion plan is re-evaluated to see whether it still intersects with the collision trajectory (Fig. \ref{fig:decision} (f)). If it does, the planner reuses the previous motion plan. In the case that the final destination does not intersect the collision trajectory, we replan the intervention trajectory. To replan the trajectory seamlessly, we attempt to reuse the same intervening body linkage of the robot computed from the previous motion planning process (Fig. \ref{fig:decision} (g)). If the reachable volume of that linkage intersects the collision trajectory, we build a new motion plan to reach one of the configurations in the volume. Otherwise, we replan from scratch the entire task.

If the current state violates the end-effector task (Fig. \ref{fig:decision} (c)), we search within the unconstrained reachable volumes.
Otherwise, we search within the constrained reachable volumes. If the search fails, the blocking task fails because the robot cannot intercept the collision trajectory.

\subsection{Constrained motion planning}
\label{sec:wbmp}
In this subsection, we consider the constrained manifold imposed by the primary end-effector goal task using whole-body operational space control (WBOSC). Thanks to the task hierarchy capability of WBOSC, we can generate a motion plan compliant with higher priority tasks. The task hierarchy of WBOSC is provided by null space projections associated with higher priority tasks. An introduction to WBOSC and detailed descriptions and derivations are provided in Appendix \ref{sec:TH}.

The rank of $\Lambda_2^*$ in that appendix, which is equivalent to that of $\Lambda_2^{*\,+}$, is the same as the dimension of the task space coordinates (e.g. the end-effector position). Thus, by the definition of $\Lambda_2^*$, the rank of $J_2^* N_1^* \Phi^* N_1^{*\,T} J_2^{*\,T}$ emerges from the following model.
\begin{align}
\begin{split}
\Lambda_2^* &= J_2^* N_1^* \Phi^* N_1^{*\,T} J_2^{*\,T} \\
			&= J_2 \overline{UN_c} N_1^* UN_c A^{-1} \left(UN_c\right)^T N_1^{*\,T} J_2^{*\,T}\\
&= J_2 N_1 A^{-1} N_1^T J_2^T \label{eq:l2}
\end{split}
\end{align}
where $N_1$ is the null space projection of the higher priority task into the lower priority task, and is defined as follows.
\begin{align}
	N_1 &= I - A^{-1} N_c^T J_1^T \left(J_1 A^{-1} N_c^T J_1^T\right)^+ J_1 N_c \label{eq:n1}
\end{align}
Because $A$ is a positive definite matrix, we only check the rank of $J_2 N_1$ to identify the dimension of the lower priority task. The rank of $N_1$ is determined from Eq. (\ref{eq:n1}). Taking the singular value decomposition of $A^{-1/2}N_c^T J_1^T$, Eq. (\ref{eq:n1}) can be expressed as follows.
\begin{align}
\begin{split}
N_1 &= I - A^{-1/2} \mathcal{U} \Sigma \mathcal{V}^T \left( \mathcal{V} \Sigma^T \Sigma \mathcal{V}^T \right)^+ \mathcal{V} \Sigma \mathcal{U}^T A^{1/2} \\
&= A^{-1/2} \left( I - \mathcal{U} \Sigma \cancel{\mathcal{V}^T \mathcal{V}} \Sigma^+ \Sigma^{+\,T} \cancel{\mathcal{V}^T \mathcal{V}} \Sigma \mathcal{U}^T \right) A^{1/2} \\
&= A^{-1/2} \left( I - \mathcal{U} \Sigma \Sigma^+ \Sigma^{+\,T} \Sigma \mathcal{U}^T \right) A^{1/2} \label{eq:n1_2}
\end{split}
\end{align}
If the rank of the diagonal matrix, $\Sigma \in \mathbb{R}^{r \times n}$ is $r < n$,  then $\Sigma \Sigma^+ \Sigma^{+\,T} \Sigma = \begin{pmatrix} I_{r \times r} & 0 \\ 0 & 0 \end{pmatrix}$ and Eq. (\ref{eq:n1_2}) becomes the following.
\begin{align}
\begin{split}
N_1 &= A^{-1/2} \left( I - \mathcal{U} \begin{pmatrix} I_{ r \times r} & 0 \\ 0 & 0 \end{pmatrix} \mathcal{U}^T  \right) A^{1/2} \\
&= A^{-1/2} \mathcal{U} \begin{pmatrix} 0 & 0 \\ 0 & I_{ (n-r) \times (n-r)} \end{pmatrix} \mathcal{U}^T  A^{1/2}
\end{split}
\end{align}
Thus, the rank of $N_1$ is determined by the robot's number of generalized coordinates minus the rank of $J_1 N_c$.

The constrained manifold can be derived by substituting Eq. (\ref{eq:l2}) into Eq. (\ref{eq:x2}) as follows.
\begin{align}
\ddot{x}_2 &= \Lambda_2^{*\,+} \Lambda_2^* \ddot{x}_{2,des} + \tau_1^{\prime \prime} \label{eq:x2u}\\
\intertext{ Because $\Lambda_2^*$ is a symmetric matrix, we can take its singular value decomposition and express it as $\mathcal{U} \Sigma \mathcal{U}^T$. Then, $\ddot{x}_2$ can be derived as follows}
\begin{split}
\ddot{x}_2&= \mathcal{U} \Sigma^+ \cancel{\mathcal{U}^T \mathcal{U}} \Sigma \mathcal{U}^T \ddot{x}_{2\,des} + \tau_1^{\prime \prime}\\
&= \mathcal{U} \begin{pmatrix} I_{r_2 \times r_2} & 0 \\ 0 & 0 \end{pmatrix} \mathcal{U}^T \ddot{x}_{2\,des} + \tau_1^{\prime \prime} \label{eq:x2u2}
\end{split}
\end{align}
where $r_2$ is the rank of $\Lambda_2^\ast$. In this equation, $\mathcal{U}$ consists of singular vectors, $\begin{pmatrix} u_1 & u_2 & \cdots & u_n \end{pmatrix}$ and $\mathcal{U}^T \ddot{x}_{2\,des}$ can be defined as an arbitrary vector, $z = \begin{pmatrix} z_1 & z_2 & \cdots & z_n \end{pmatrix}^T $. Then, $\ddot{x}_2$ can be expressed as the weighted sum of the independent basis as follows
\begin{align}
\ddot{x}_2 &= z_1 u_1 + z_2 u_2 + \cdots + z_{r_2} u_{r_2} + \tau_1^{\prime \prime} \label{eq:sv}
\end{align}
Therefore, the singular vectors of the task space mass matrix, $\Lambda_2^*$ are the basis of the constraint manifold considering the null space of higher priority tasks.

From the previous section, the basis of the lower priority task has been derived. With this basis, we can generate the $Project$ function shown in Algorithm 1. The arguments of that function are the current robot configuration and the displacement of the joints. To make the displacement compliant with the end-effector task constraint, it needs to be projected into the null space of the end-effector based on Eq. (\ref{eq:x2u}). $\Lambda_2^\ast$ is the dynamic term as a function of $q$, such that the projected displacement shown in Algorithm 1 takes the form.
\begin{align}
\ddot{x}_2 = \Lambda_2^{\ast\, +}(q) \Lambda_2^\ast(q) \left( x_{disp} - \tau^\prime \right) + \tau^\prime
\end{align}




\section{Experiments}
\subsection{Experimental setup}
\begin{figure}[t]\centering
\ifdefined\JOURNAL
\includegraphics[width=\linewidth, clip=false ] {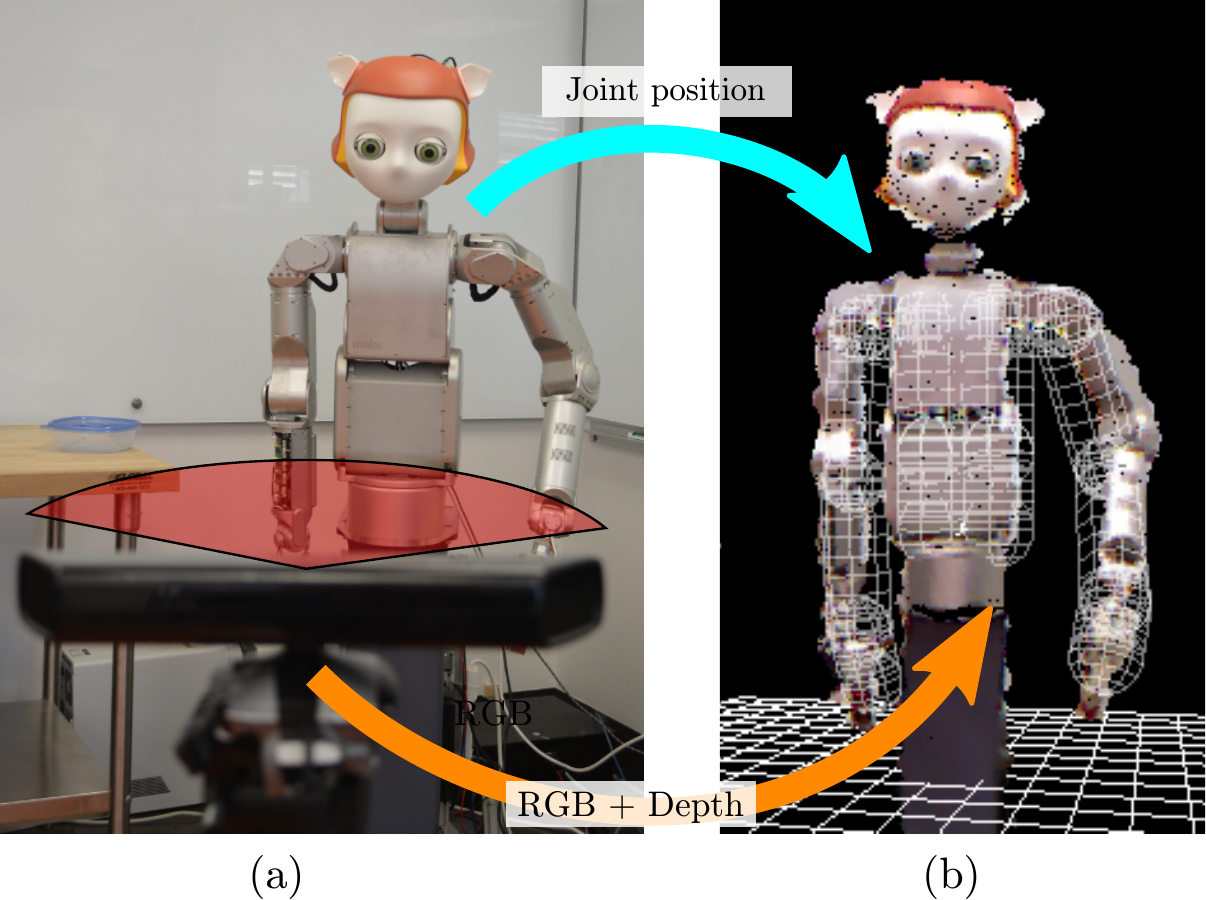}
\else
\includegraphics[width=\linewidth, clip=false ] {setup2.pdf}
\fi
\caption
[The experimental setup]
{{\bf The experimental setup}
consists of an external Kinect camera and the Dreamer robot (a) which consists of an upper-body humanoid robot. Dreamer consists of two 7-DoF arms a 3-DoF torso, the latter including two forward bending coupled joints and a vertical rotation. Each robot arm consists of 7 revolute joints. We approximate the forward bending torso with two capsules. The simplified stick figure for the robot (only considering the torso and left arm) is shown in (b). 
The transformation between the robot coordinates and the Kinect coordinates is calculated beforehand, so all the RGB points from Kinect can be represented in the robot's coordinates as shown in (b). 
} \label{fig:setup} 
\end{figure}
To validate the proposed methods, we conduct collision intervention experiments using an upper-body humanoid robot, Dreamer, shown in Fig. \ref{fig:setup}. Detailed information about the robot can be found in \cite{sentis2012}.

To sense its environment, the robot uses a Kinect sensor to detect external objects. Objects in the robot's environment are identified via a color profile and markers. The color profile of a tennis ball is considered in advance, and object markers are set to be tracked using the Aruco computational library \cite{munoz2012}. Therefore, whenever the Kinect sensor receives a new RGB image frame with a corresponding depth image, the blocking robot can track the green tennis ball and the various registered markers such as the one shown in Fig. \ref{fig:concept}, while the locations and the velocities of the objects are estimated via Kalman filters as shown in Eq. (\ref{eq:st}). The parameters for the Kalman filter are shown in Table \ref{tab:kalman}. Given those states, the collision probability between two objects can be estimated as shown in Sec. \ref{sec:colman}.
The whole experimental setup with Dreamer and the Kinect sensor is shown in Fig. \ref{fig:setup}.
\begin{table}
\caption
[The Kalman filter parameters for object tracking]
{\rm The Kalman filter terms and parameters for object tracking}\label{tab:kalman}
\begin{center}
{\renewcommand{\arraystretch}{1.5}
\begin{tabular}{|c|c|c|c|}
\hline
Name & Description & Definition/Value & Metric \\
\hline
\hline
$\xi_p^i$ & \makecell{Random variable of \\state estimation noise} & $ \left( \mathbf{0},\  \begin{pmatrix} \Sigma_d & 0 \\ 0 & \Sigma_\alpha \end{pmatrix} \right)$ & 
\\
\hline
$\xi_s^i$ & \makecell{Random variable of \\sensor noise} & $\left( \mathbf{0},\  \Sigma_s \right)$ & \\
\hline
$\hat{P}_m^i$ & Predicted state& $A_p \hat{P}_{m-1}$ & \\
\hline
$\hat{\Phi}_m^i$ & Predicted covariance& $A_p \Phi_{m-1} A_p^T + \Sigma_w^i$ & \\
\hline
$S_m^i$ & Innovation covariance & $C_p \Phi_{m|m-1}^i C_p^T + \Sigma_s^i$ & \\
\hline
$K_m^i$ & Kalman gain & $\Phi_{m|m-1}^i C_p^T {S_m^i}^{-1}$ & \\
\hline
$\Delta t$ & Sampling time & 33 & \rm msec \\
\hline
$\Sigma_d$ & \makecell{Covariance of velocity\\ disturbance} & diag(0.01,0.01,0.01) & $\rm \left( m / s \right)^2 $ \\
\hline
$\Sigma_\alpha$ & \makecell{Covariance of \\acceleration} & diag(1.5, 1.5, 1.5) & \rm $\rm \left( m / s^2 \right)^2$ \\
\hline
$\Sigma_s$ & Position sensor noise & diag(0.01,0.01,0.01) & $\rm m^2 $ \\
\hline
\end{tabular}
}
\end{center}
\end{table}

The blocking robot, Dreamer, is set to perform a primary goal task in which the end-effector moves and remains at a stationary goal position. Therefore collision intervention tasks to stop external objects are attempted to be performed under the null space of the primary task when possible. The end-effector's stationary goal is shown as a red ball in Fig. \ref{fig:occu}-(a). Before a collision intervention task takes place, the robot generates the constrained and unconstrained probabilistic roadmaps using 10,000 samples, and the reachable volumes of the shoulder and arm linkages corresponding to the roadmaps are also generated, as shown in Fig. \ref{fig:occu}-(a) and (b). The volume of the operating environment around the blocking robot is $2 \rm m \times 2 \rm m \times 2 \rm m = 8 \rm m^3$, and we segment it via the Octree algorithm with 6 levels of depth. Therefore, we consider 8 reachable volumes corresponding to the constrained/unconstrained roadmaps and the four robot linkages. The dimension of each cube in the Octree is $3 \rm cm \times 3 \rm cm \times 3 \rm cm$. The generated reachable volumes of the constrained roadmap for the 4 linkages are shown in Fig. \ref{fig:occu}. If the object trajectory overlaps any of the constrained volumes, the blocking robot is able to simultaneously stop the external object's movement while keeping the end-effector on its goal position. The search for this process is conducted based on the decision policy shown in Fig. \ref{fig:decision}.
\begin{figure}[ht]\centering
\ifdefined\JOURNAL
\includegraphics[width=1\linewidth] {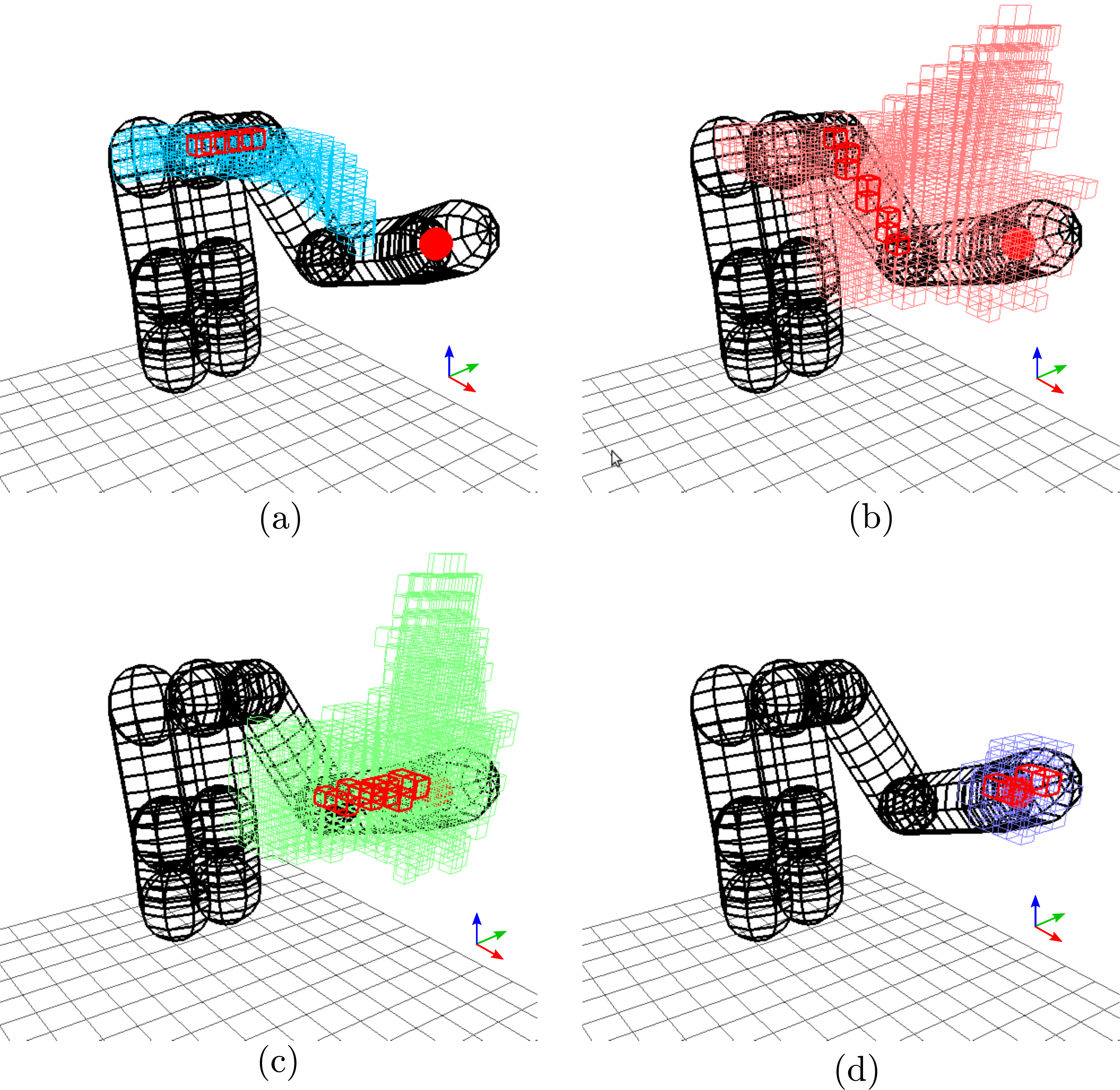}
\else
\includegraphics[width=0.8\linewidth] {occu.pdf}
\fi
\caption
[ The reachable volumes of the links corresponding to the constrained roadmap]
{{\bf The reachable volumes of the linkages corresponding to the constrained roadmap} are generated for the occupancies of the shoulder, the upper arm, the lower arm, and the end-effector. 
} \label{fig:occu} 
\end{figure}

\subsection{Constrained collision intervention}
\begin{figure}[t]\centering
\ifdefined\JOURNAL
\includegraphics[width=\linewidth] {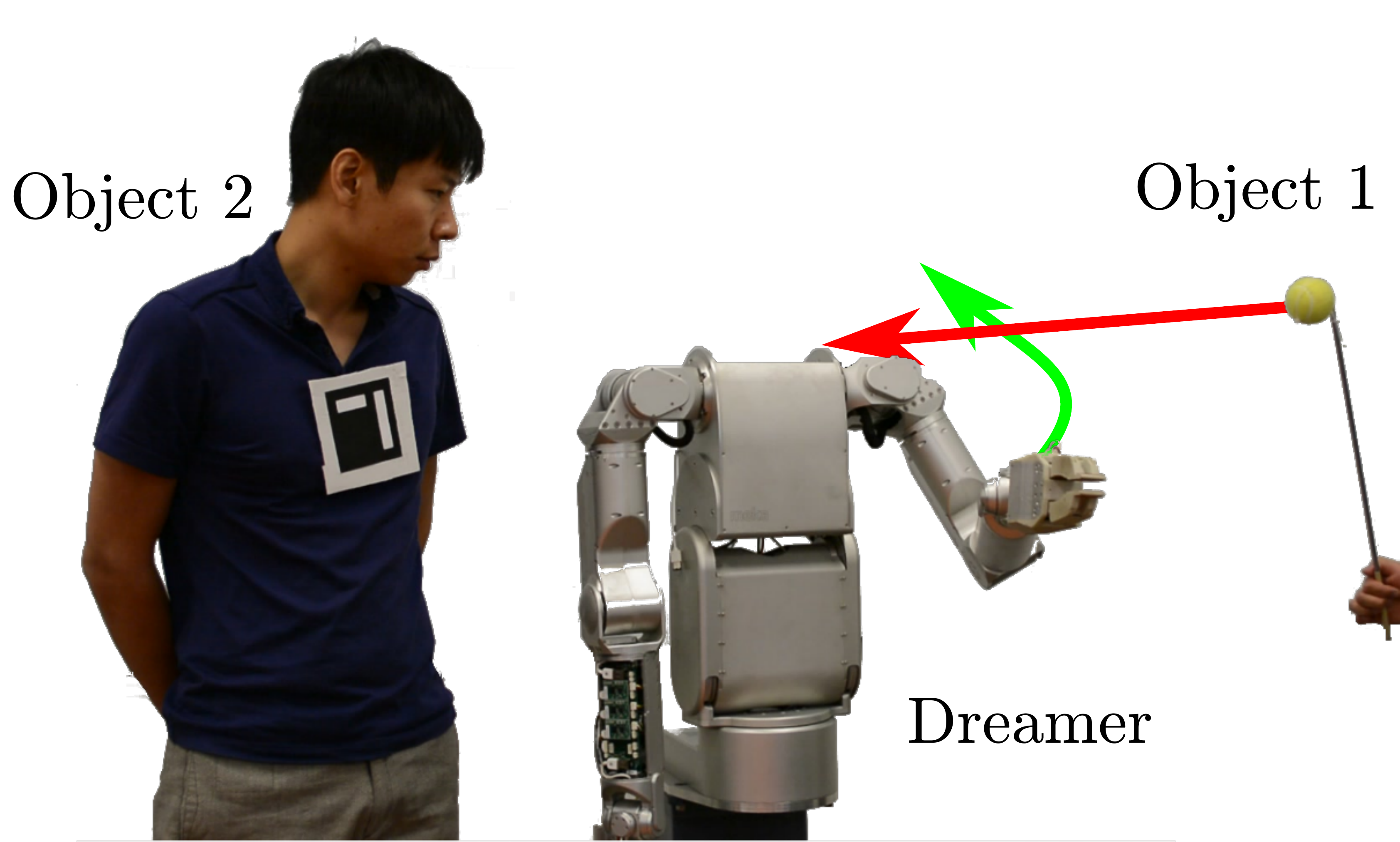}
\else
\includegraphics[width=0.8\linewidth] {concept2.pdf}
\fi
\caption
[ Concept image of the experiments]
{{\bf Concept image of the experiment}. The blocking robot intercepts object 1 if it is determined that object 1 is set to collide with object 2. We assume that object 2 is a placeholder for a human operator who needs to be protected against object 1, which is a ball. The blocking robot tracks the marker and the green ball using Aruco and the OpenCV library. During the tracking process, the positions and velocities are estimated via Kalman filter. The Kinect sensor updates the input video sequence every 33msec and so does the Kalman filter.
} \label{fig:concept} 
\end{figure}
In this experiment, the object with the marker represents a human subject working nearby the blocking robot, and the ball represents a moving object harmful to the subject. The objective of the blocking robot is to 1) observe the human and the ball, 2) determine whether these objects will collide with each other, and 3) stop potential collisions between these two objects by using one of its body parts. This scene is illustrated in Fig. \ref{fig:concept}. Object 1 is the ball, which approaches object 2, the human subject. The movement of the three objects including the blocking robot is estimated, and the probability that the external objects collide with each other is predicted based on Eq. (\ref{eq:pac}). To determine whether there will be a collision or not, the threshold values defined in Fig. \ref{fig:cum_col} needs to be set. In this experiment, we use $\eta = 0.5$ and $t_c = 4$, which means there will be a collision if the cumulative collision probability is 0.5 or above it and within a 5 seconds prediction horizon. If the collision probability exceeds the threshold, the robot searches its reachable volumes for postures intercepting the trajectory of the external objects as shown in Fig. \ref{fig:int_l}. According to the search policy shown in Fig. \ref{fig:decision}, the blocking robot searches the constrained reachable volumes first. If it fails to find a feasible solution, it searches the unconstrained reachable volumes. If the robot finds an intercepting final pose, it will generate a motion path from the current pose to the intercepting pose given the PRM plan.

Three separate collision intervention processes are shown in Figs. \ref{fig:col1} and \ref{fig:col3}. In all of these cases the first set of images depict the robot determining that objects 1 and 2 will collide against each other, and switching its state to the intervention mode. In the intervention case shown in Fig. \ref{fig:col1}-(a), the blocking robot intervenes using its upper arm, while in Fig. \ref{fig:col1}-(d) it determines that the best body part is using its lower arm. In addition, Fig. \ref{fig:col3} shows an intervention process using the robot's shoulder. Reachable volume data analysis is further shown in Fig. \ref{fig:occu1} for the first two experiments. 

\begin{figure}[t]\centering
\ifdefined\JOURNAL
\includegraphics[width=1.0\linewidth, clip=false ] {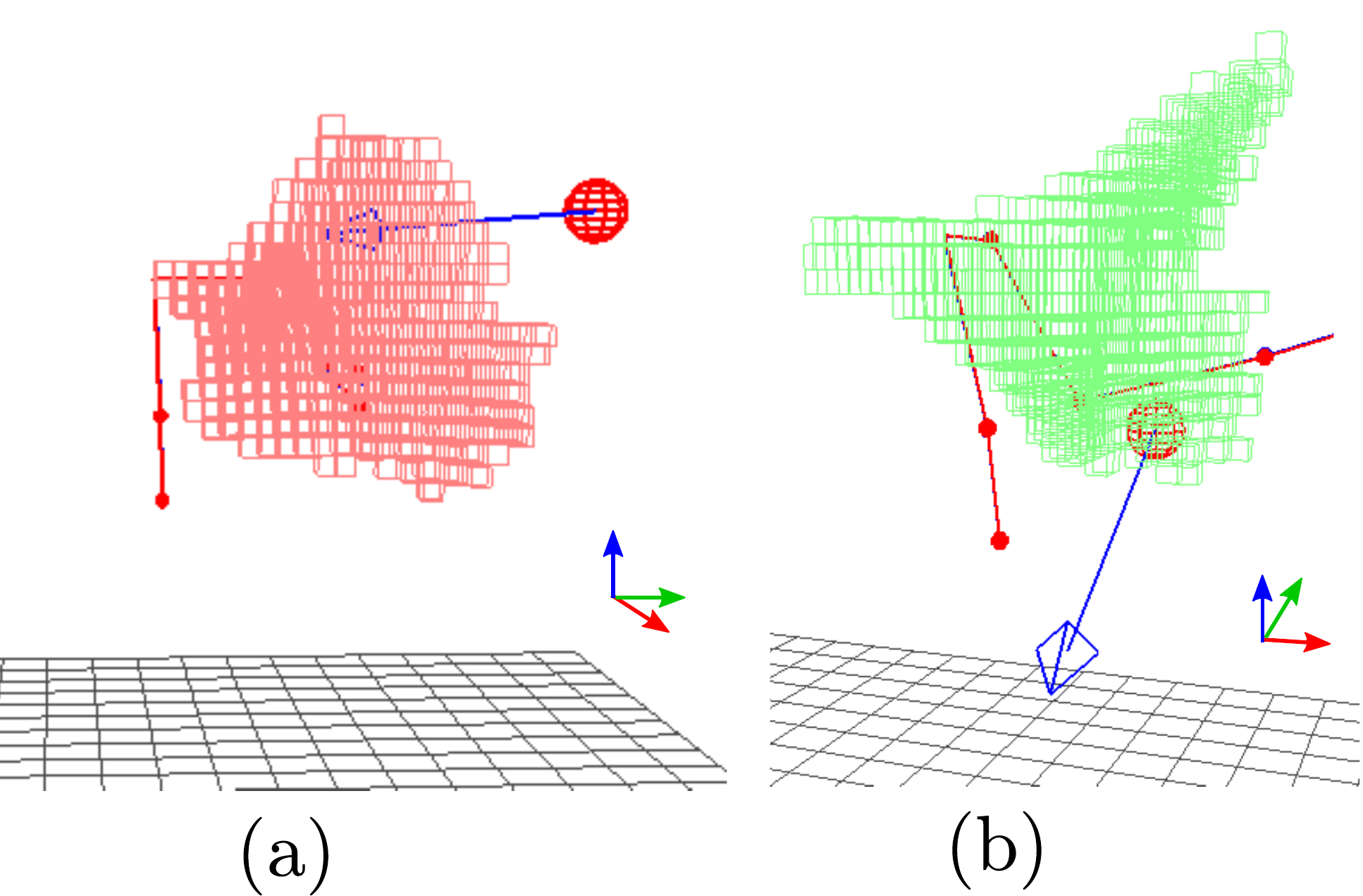}
\else
\includegraphics[width=0.7\linewidth, clip=false ] {occu1.pdf}
\fi
\caption
[ Reachable volume during collision intervention]
{{\bf Reachable volume during collision intervention}. When object 1 approaches object 2 in Fig. \ref{fig:col1}, the reachable volume of the upper arm overlaps the ball trajectory shown as a blue arrow in (a). In contrast, the estimated trajectory of the ball shown in Fig. \ref{fig:col1}-(a) overlaps with the lower arm as shown in (b).} \label{fig:occu1} 
\end{figure}

\begin{figure*}[p]\centering
\includegraphics[width=0.9\linewidth, clip=false ] {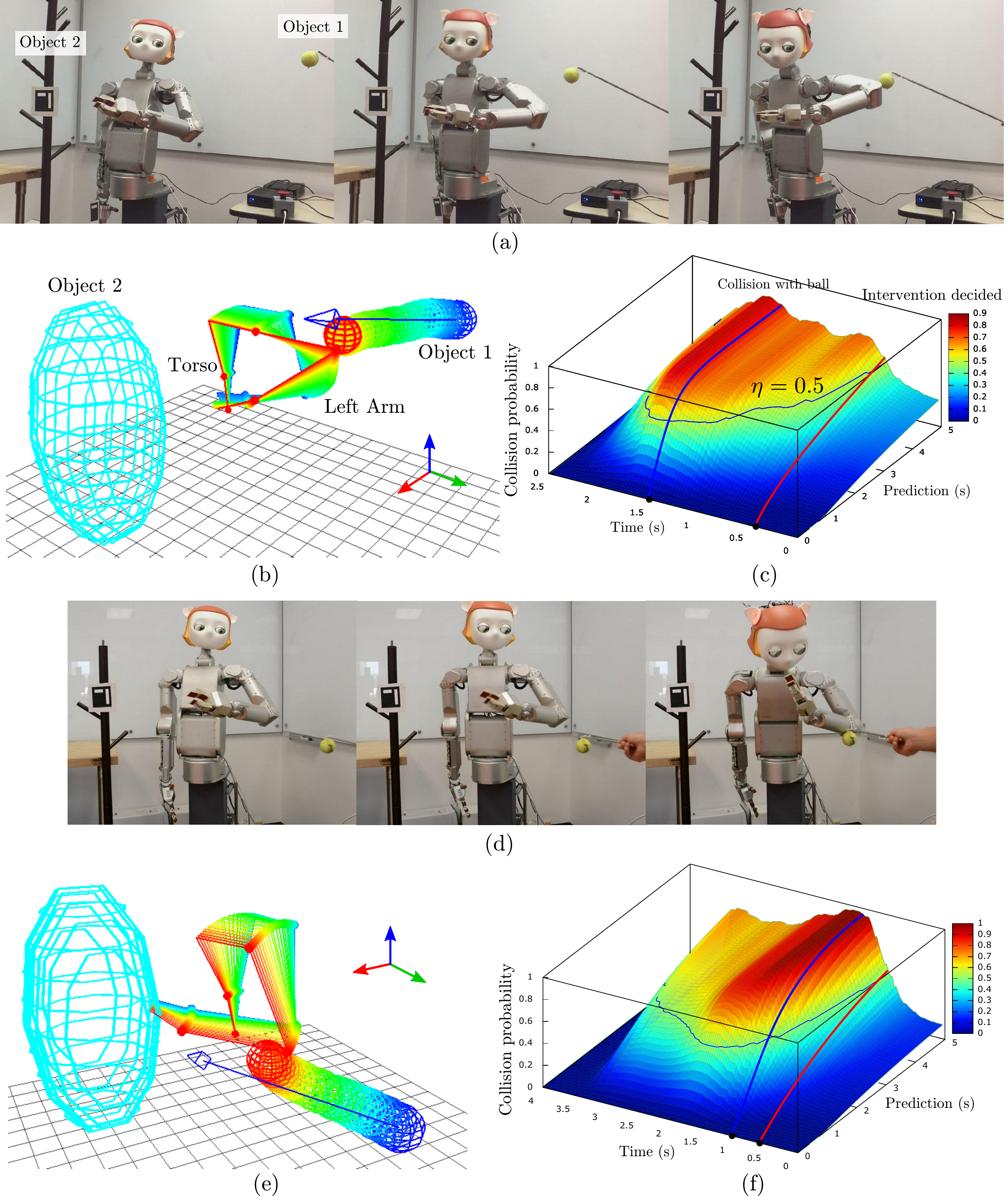}
\caption
{{\bf Collision intervention using the robot's upper and lower arm}. The robot generates a motion plan which makes its upper arm stop object 1 from colliding with object 2 (a). Subfigure (b) shows the posture and ball trajectories. Subfigure (d) shows the predicted collision probability at each time frame. At time 0.5, the probability exceeds the threshold value within 5 seconds, and therefore the blocking robot chooses to stop the likely collision. The blue arrow corresponds to the predicted object motion. The robot's lower arm blocks object 1 from colliding with object 2 (d), and its trajectory and collision probability are shown in (e) and (f), respectively.
} \label{fig:col1} 
\end{figure*}

\subsection{Unconstrained collision intervention}
\begin{figure}[t]\centering
\begin{minipage}[t]{0.47\linewidth}
\includegraphics[width=\linewidth, clip=false ] {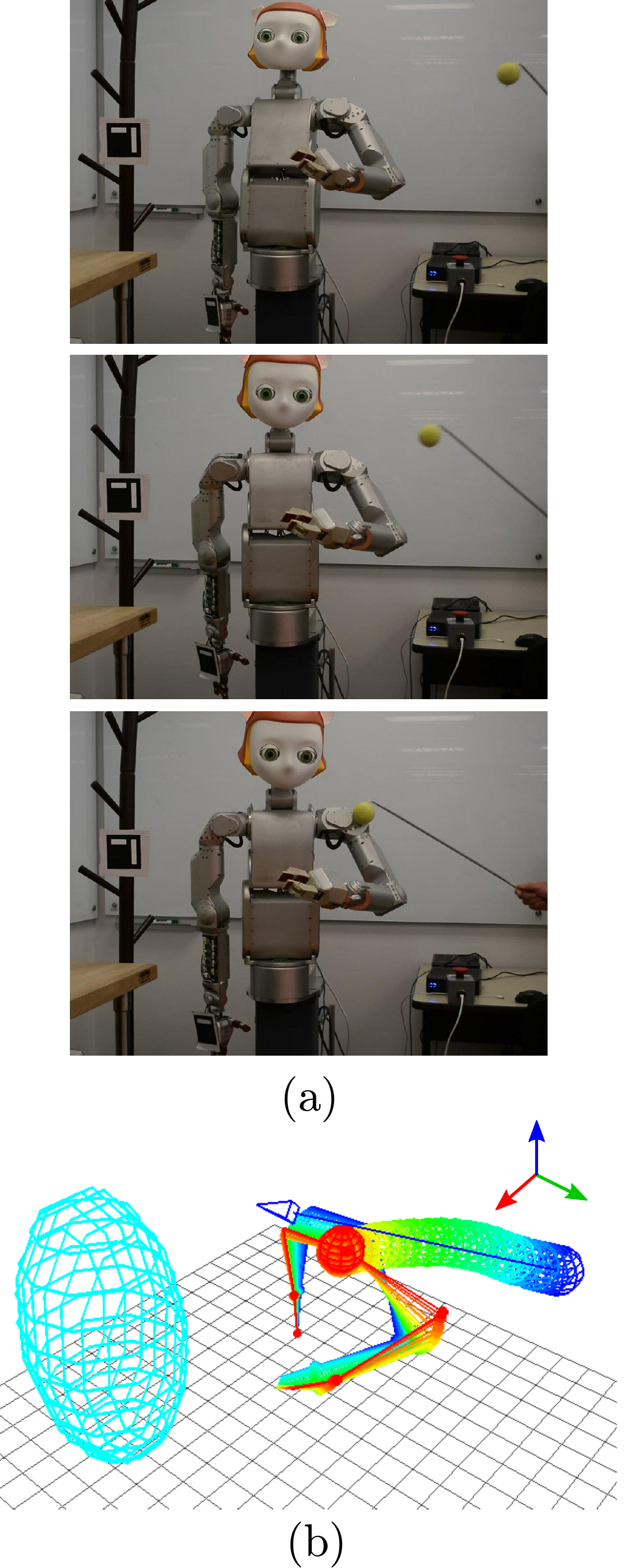}
\caption
[ Collision intervention using the robot's shoulder]
{{\bf Collision intervention using the robot's shoulder}. The reachable volume of the robot's shoulder overlaps with the estimated ball trajectory and as a result the collision intervention process is chosen to occur using the shoulder.
} \label{fig:col3} 
\end{minipage}
\begin{minipage}[t]{0.47\linewidth}
\includegraphics[width=\linewidth, clip=false ] {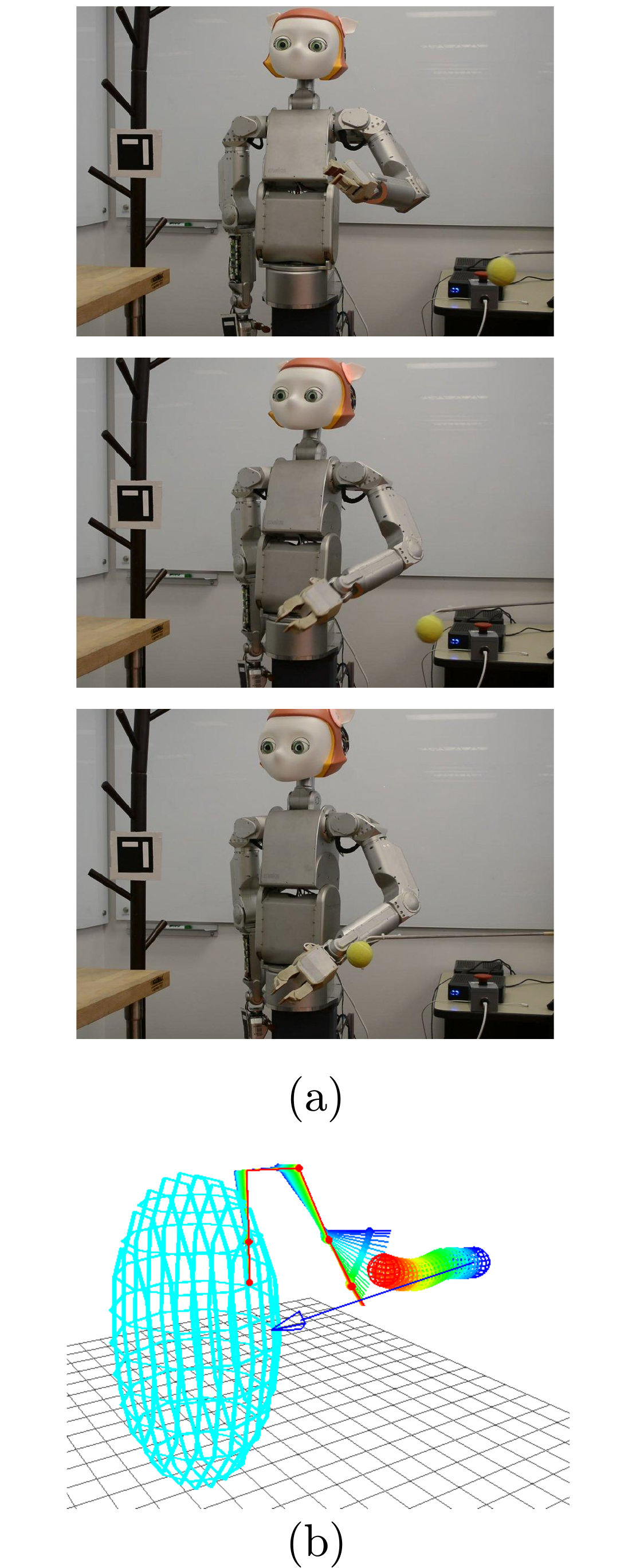}
\caption
[Unconstrained collision intervention using the robot's lower arm]
{{\bf Unconstrained collision intervention using the robot's lower arm}. Here, there is no constrained reachable volume that can be used for collision intervention. As a result the robot intervenes using its unconstrained lower arm.
} \label{fig:col4} 
\end{minipage}
\end{figure}
\begin{figure}[t]\centering
\ifdefined\JOURNAL
\includegraphics[width=1.0\linewidth, clip=false ] {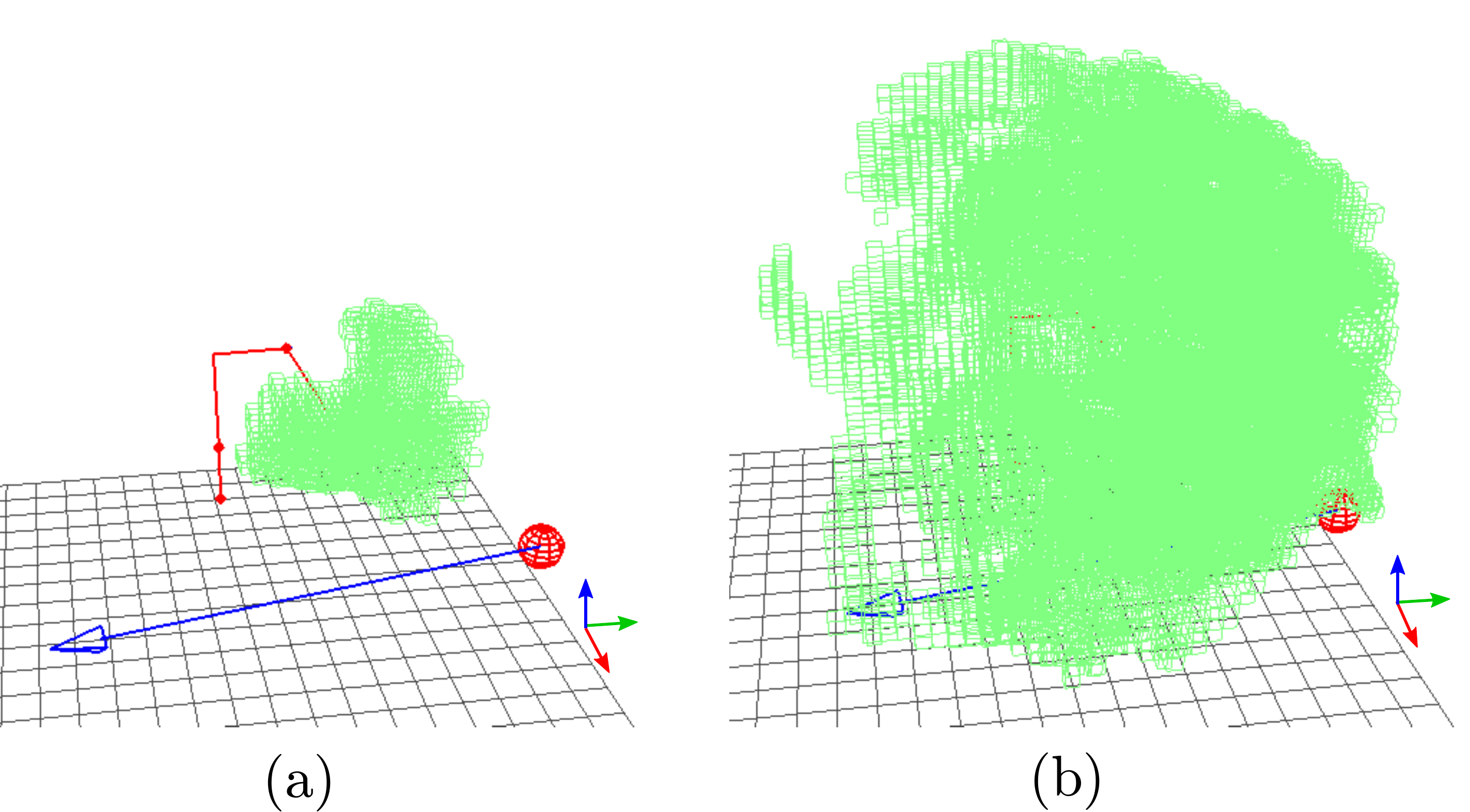}
\else
\includegraphics[width=0.8\linewidth, clip=false ] {occu2.pdf}
\fi
\caption
[Constrained and unconstrained reachable volumes]
{{\bf Constrained and unconstrained reachable volumes} are shown in subfigures (a) and (b), respectively. Those volumes correspond to the reasoning process of the experiment shown in Fig.~\ref{fig:col4}. The estimated object trajectory of Fig. \ref{fig:col4} is shown as blue arrows. Because some cubes of the unconstrained reachable volume shown in (b) overlap with the trajectory, the unconstrained lower arm is chosen to be used for intervention. } \label{fig:occu2} 
\end{figure}
If the predicted external object's trajectory does not overlap with any of the constrained reachable volumes of the blocking robot, the robot cannot intervene while satisfying the primary end-effector goal task. In the experiment shown in Fig.~\ref{fig:col4}, the blocking robot is allowed to violate the goal task. Fig. \ref{fig:occu2} shows the constrained and unconstrained reachable volumes corresponding to that experiment, which justify the decision to violate the primary goal task. 
In the first image sequence of Fig. \ref{fig:col4}, object 1 approaches object 2, and when collision probability exceeds the designed threshold, the robot decides to intervene.

\subsection{Concept experiment blocking and external robot}
Fig. \ref{fig:trikey_dreamer} shows experiments involving interventions of the upper-body humanoid robot Dreamer to stop a ground mobile robot, Trikey from colliding with a person. While the mobile robot approaches the standing person, the blocking robot estimates the probability of an external collision. If it exceeds a set threshold, the blocking robot intercepts the collision by stopping the mobile robot. In addition, the mobile robot incorporates a control algorithm that changes direction once it senses a contact as described in \cite{Kim2015}. As a result once touched the mobile robot will move away on the opposite direction thus preventing injuring the human. This study assumes that every now and then collision avoidance on mobile robots will fail to work and other means of safety could potentially be beneficial like the one considered here.
\begin{figure}[t]\centering
\ifdefined\JOURNAL
\includegraphics[width=\linewidth] {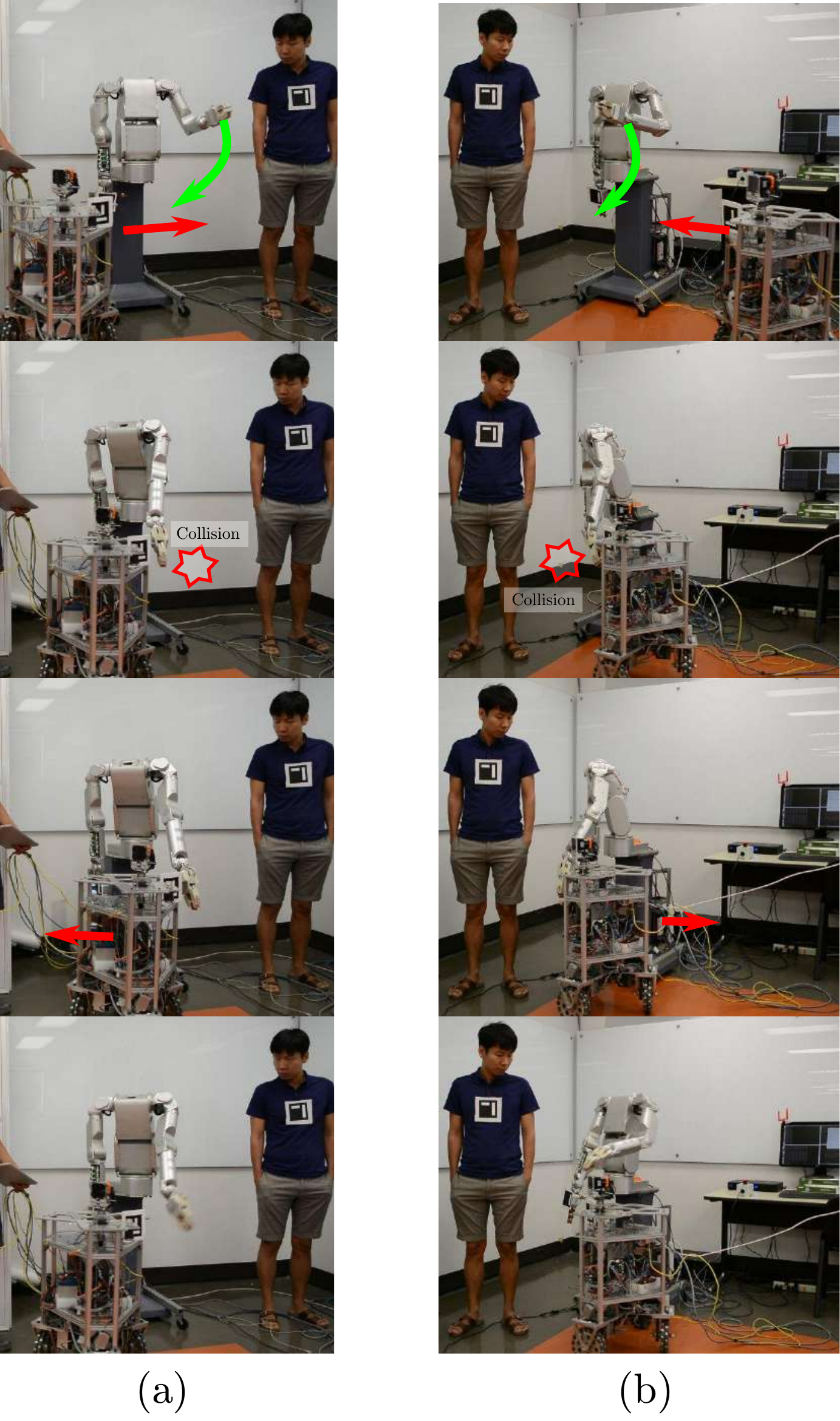}
\else
\includegraphics[width=0.7\linewidth] {trikey_dreamer.pdf}
\fi
\caption
[ Proof of concept experiment between two robots]
{{\bf A robot blocking another robot}. A ground mobile robot dangerously approaches a human subject. The blocking humanoid robot intervenes to stop the likely collision to prevent injury to the person.
} \label{fig:trikey_dreamer} 
\end{figure}

\section{Conclusion}
In this paper, we investigate methods for safety by blocking potential collisions between external objects or robots and people. We estimate collision probabilities and compared against set thresholds. Secondly, we generate motion plans to block likely collisions using an upper body humanoid robot. For intervention, we consider any body part of the entire robot as candidates for blocking the collision. We also assume that the robot is already engaged in a primary goal task using its end-effector, and we utilize the redundant degrees of freedom to stop external collisions when possible. We computationally build constrained reachable volumes for all robot's body parts and interconnect  corresponding postures via PRM planning. We then check whether the reachable volumes overlap with the collision trajectories. If any of the reachable volumes does overlap, a motion plan is generated to accomplish the intervention. Otherwise, the blocking robot searches for overlapping postures within the unconstrained reachable volumes. These methods are implemented in a a set of robotic systems involving collisions between a ball, a mobile robot, and a human subject. As future extensions, fast visual sensors and time-optimal controllers could be adopted to reduce the response times. Also, our task planner could be improved by considering not only the collision probability but also the severity of collisions, and the importance of each object in the environment.

\appendices
\section{Whole-body operational space control}
\label{sec:TH}
A robot's dynamic equation incorporating contacts constraints has the following expression \cite{kim2013}.
\begin{align}
A \ddot{q} + N_c^T bg = \left(U N_c\right)^T \tau
\end{align}
where $A$, $q$, $N_c$, $bg$, $U$, and $\tau$ are the mass matrix, the generalized robot coordinates, the constraint null space projection associated with contacts, Coriolis-centrifugal-gravity terms, the selection matrix of actuated joints, and torque/force input to the robot, respectively. According to the WBOSC framework, the torque input for the highest priority task is computed as follows.
\begin{align}
\begin{split}
A \ddot{q} + N_c^T bg = &
\left(U N_c\right)^T \left( J_1^{*T}\Lambda_1^*\ddot{x}_{1,des} + N_1^{*\,T} \tau_2 \right)\\
& + \left(U N_c\right)^T \left( \overline{UN_c}^T bg\right)
\end{split}
\\
	\intertext{where $J_1^*$, $\Lambda_1^*$, $N_1^*$, $\overline{(\cdot)}$, and $\tau_2$ are the Jacobian matrix of the highest priority task, the highest priority task's mass matrix, the null space of the task, the dynamically consistent pseudo-inverse of $(\cdot)$, and the torque input of lower priority tasks, respectively. If the robot is fully controllable under contact constraints, which means $\overline{UN_c} UN_c = N_c$, then $bg$ can be canceled out as follows.}
A \ddot{q}  =& \left(U N_c\right)^T \left( J_1^{*T}\Lambda_1^*\ddot{x}_{1,des} + N_1^{*\,T} \tau_2 \right) \label{eq:dyn2}
\end{align}
Lower priority tasks should be compliant with higher priority tasks, so the lower priority tasks are constrained. In Eq. (\ref{eq:dyn2}), the lower priority task input $\tau_2$ is projected into the null space of the higher priority task by left-multiplying by $N_1^*$. $N_1^*$ is defined as follows.
\begin{align}
N_1^* &= I - \overline{J_1^*} J_1^*
\end{align}

By left-multiplying $J_2 A^{-1}$ by Eq. (\ref{eq:dyn2}), the constraint dynamic equation becomes
\begin{align}
\begin{split}
	\ddot{x}_2 + \dot{J}_2 \dot{q}  &= J_2 A^{-1}\left(U N_c\right)^T \left( \tau_1 + N_1^{*\,T} J_2^{*\,T}\Lambda_2^* \ddot{x}_{2,des} \right) \\
&= \tau_1^\prime + J_2^* \Phi^* N_1^{*\,T} J_2^{*\,T} \Lambda_2^* \ddot{x}_{2,des}\\
&= \Lambda_2^{*\,+} \Lambda_2^* \ddot{x}_{2,des} + \tau_1^\prime \label{eq:x2}
\end{split}
\end{align}
If the lower priority task's mass matrix, $\Lambda_2^*$ is full rank, the task is fully controllable and $\ddot{x}_2$ equals the desired task acceleration, $\ddot{x}_{2,des}$.

\section{Collision probability between two objects}
\label{sec:pb}
\subsection{Probability of collision}
\label{sec:a_pc}
For easy understanding, a one-dimensional collision between two objects labeled as $i$ and $j$ is considered first. Let's assume that there are two one-dimensional objects consisting of line segments with length $l_i$ and $l_j$, and their center positions are distributed with probability densities, $\mathbf{P}_o^i$ and $\mathbf{P}_o^j$. Then, the probability of a collision between them is described as follows.
\begin{align}
\begin{split}
\mathbf{p}_{ic}^{ij} =
&\int_{-\infty}^{\infty}
\int_{-\infty}^{\infty}
f_{ij}(x_i,\, x_j) \mathbf{P}_o^i (x_i) \, \mathbf{P}_o^j(x_j) dx_j\, dx_i
\\
=&\int_{-\infty}^{\infty}
\mathbf{P}_o^i (x_i) 
\left(\int_{x_i-w}^{x_i+w}
	 \mathbf{P}_o^j(x_j) dx_j\, \right)dx_i\\
=&\int_{-\infty}^{\infty}
\mathbf{P}_o^i (x_i) 
\left(\int_{-w}^{+w}
	 \mathbf{P}_o^j(x_i+x) dx\, \right)dx_i \label{eq:a_pc}
\end{split}
\end{align}
where $x \triangleq x_j - x_i$ is the relative position of $x_j$ with respect to $x_i$, and $f_{ij}$ is the predicate function which returns $1$ if the two objects positions $x_i$ and $x_j$ end up colliding as follows.
\begin{align}
f_{ij}(x_i, x_j)
\bigg\{	\begin{matrix}
1 & -w \le x_i - x_j  \le  w\\
0 & \rm otherwise
\end{matrix}\label{eq:fij}
\end{align}
where  $w \triangleq l_i + l_j$.

Using the Fubini theorem, the two integrals commute. 
\begin{flalign}
\begin{split}
=&
\int_{-w}^{+w}
\left(
\int_{-\infty}^{\infty}
\mathbf{P}_o^i (x_i) 
	 \mathbf{P}_o^j(x_i+x) 
	 dx_i\,
	 \right)
	dx \\
=&
\int_{-w}^{+w}
\left(
\int_{-\infty}^{\infty}
\mathbf{P}_o^i (x_i) 
	 \mathbf{P}_o^j\left(-(-x-x_i) \right) 
	 dx_i\,
	 \right)
	dx 
\end{split}
\end{flalign}
Defining a new function, ${\mathbf{P}_o^j}^\prime$ such that ${\mathbf{P}_o^j}^\prime(x) \triangleq \mathbf{P}_o^j(-x)$, the probability becomes the integral of the convolution as follows.
\begin{flalign}
=&
\int_{-w}^{+w}
\Big(
\mathbf{P}_o^i(-x) \ast {\mathbf{P}_o^j}^\prime(-x) 
\Big)
dx 
\end{flalign}
In the case of the one-dimensional scenario, the shape of objects are line segments, which are symmetric, so the integration of their Minkowski sum is also symmetric.
\begin{flalign}
=&
\int_{-w}^{+w}
\Big(
\mathbf{P}_o^i(x) \ast {\mathbf{P}_o^j}^\prime(x) 
\Big)
dx 
\end{flalign}
The convolution of two normal distributions are also a normal distribution, and its mean and covariance are derived from those of the two normal distributions. So, provided that $\mathbf{P}_o^i$ and $\mathbf{P}_o^j$ are assumed to be normal distribution with the following properties, 
\begin{align}
\begin{split}
\mathbf{P}_o^i(x_i) = \mathcal{N}\left( \mu_i,\ \sigma_i^2 \right) \\
\mathbf{P}_o^j(x_j) = \mathcal{N}\left( \mu_j,\ \sigma_j^2 \right)
\end{split},
\end{align}
the convolution of the two normal distributions becomes another normal distribution, $\mathbf{P}_{conv}$ as follows.
\begin{align}
\mathbf{P}_{conv}(x) &= \mathbf{P}_o^i(x) \ast {\mathbf{P}_o^j}^\prime(x) \nonumber\\
&= \mathcal{N} \left( \mu_i - \mu_j,\ \sigma_i^2 + \sigma_j^2 \right)
\end{align}
Therefore, the probability of collision between the two objects in one-dimensional space is the cumulative density function of the new normal distribution, and the instantaneous collision probability between the two objects, $\mathbf{p}_{ic}^{ij}$, becomes a cumulative density function of a normal distribution as follows.
\begin{align}
\mathbf{p}_{ic}^{ij} &= \int_{-\omega}^{\omega} \mathbf{P}_{conv} \left( x \right) dx
\end{align}
Depending on the distance of the two normal distributions and the size of the objects, there are two ways to express the collision probability using the error function, $erf$.
\begin{align}
\mathbf{p}_{ic}^{ij} =
\begin{cases}
\frac{1}{2}\left( erf\left( x^+\right) - erf\left( x^- \right) \right)
& \left| \mu_i - \mu_j \right| > \omega \\
\frac{1}{2}\left( erf\left( x^+ \right) + erf\left( - x^- \right) \right)
& \left| \mu_i - \mu_j \right| \le \omega
\end{cases} 
\end{align}
where $x^+ \triangleq \frac{ \left| \mu_i - \mu_j \right| + \omega}{\sqrt{\sigma_i^2 + \sigma_j^2}}$ and 
 $x^- \triangleq \frac{ \left| \mu_i - \mu_j \right| - \omega}{\sqrt{\sigma_i^2 + \sigma_j^2}}$.

To extend collisions in one-dimensional space to 3D space, the overlapping predicate function of Eq. (\ref{eq:fij}) also needs to be extended. If it is assumed that the objects are rigid bodies, then the predicate depends only on their relative position and we can define another predicate function associated with the relative position, $F_{ij}$ as follows.
\begin{align}
F_{ij}(x) \triangleq f \left( 0,\, x \right)
\label{eq:Fij}
\end{align}
Because the predicate only depends on the relative position of the two objects, it has the following property.
\begin{align}
\forall x_i.\, 
F_{ij}(x) = f \left( x_i,\, x_i + x \right)
\label{eq:Fij2}
\end{align}
Also, we can define a Minkowski sum of the two objects as $\mathcal{B}$, and the integral of an arbitrary function over the region has the following property.
\begin{align}
\int_{\mathcal{B}} g(x) dx = \int_{-\infty}^{\infty} 
F_{ij}(x)\, g(x) dx
\end{align}
Then, the one-dimensional collision probability in Eq. (\ref{eq:a_pc}) is extended to the 3D case as follows.
\begin{align}
\begin{split}
&\int_{-\infty}^{\infty}
\mathbf{P}_o^i (x_i) 
\left(\int_{\mathcal{B}}
	 \mathbf{P}_o^j(x_i+x) dx\, \right)dx_i \\
=&
\int_{\mathcal{B}}
\Big(
\mathbf{P}_o^i(-x) \ast {\mathbf{P}_o^j}^\prime(-x) 
\Big)
dx
\end{split}
\end{align}
where $\mathbf{P}_o^i$ and $\mathbf{P}_o^j$ are joint probability density functions in 3D with the following properties.
\begin{align}
\begin{split}
\mathbf{P}_o^i(x_i) = \mathcal{N}\left( \mu_i,\ \Sigma_i \right) \\
\mathbf{P}_o^j(x_j) = \mathcal{N}\left( \mu_j,\ \Sigma_j \right)
\end{split}
\end{align}
The convolution of the normal distribution in 3D space is also a joint normal distribution. So, the collision probability in 3D is also a cumulative probability function of a normal distribution as follows.
\begin{align}
\mathbf{p}_{ic}^{ij} &= \int_{\mathcal{B}} \mathbf{P}_{conv} \left( x \right) dx
\end{align}
where $\mathbf{P}_{conv}$ is the convolution of the two normal distributions which is another normal distribution with the following property.
\begin{align}
\mathbf{P}_{conv} &\sim \mathcal{N} \left( 
		\mu_i - \mu_j,\ 
		\Sigma_i + \Sigma_j
\right)
\end{align}
In this paper, we assume that all objects are spheres and non-rotating, so the Minkowski sums are always ellipsoid. Therefore, the collision probability is derived from the integration of the normal distribution around an ellipsoid. A computationally efficient integration algorithm is considered from \cite{sheil1977} to compute the collision probability.

\subsection{Conditional probability density function of collision}
\label{sec:xf}
From Eq. (\ref{eq:a_pc}), the probability that the i-th object located at $p^i$ collides with the j-th object in 3D can be described as follows.
\begin{align}
\begin{split}
\mathbf{p}_{B}^{ji} (p^i) =&
\int_{-\infty}^{\infty}
f_{ij}(p^i,\, p^j) P_o^j(p^j) dp^j \\
=&
\int_{-\infty}^{\infty}
F_{ij}(p^j- p^i) P_o^j(p^j) dp^j \\
=&
\int_{-\infty}^{\infty}
F_{ij}\big( -\left(p^i - p^j\right)\big) P_o^j(p^j) dp^j \\
=&\,
F_{ij}\left( -p^i\right) \ast P_o^j(p^i) \label{eq:conv}
\end{split}
\end{align}
where $f_{ij}$ and $F_{ij}$ are the predicate functions defined in Eqs. (\ref{eq:fij}) and (\ref{eq:Fij}). Using the probability distribution, the corresponding random variable for the first object, $X_{c,1}$ can be defined. We can assume that the probability of the second object $P_2$ is a normal distribution, but $F_{ij}$ is obviously a function that indicates the Minkowski sum of the two objects. However, based on the central limit theorem that repeated convolutions of two probability density functions converge to a normal distribution, we can assume that the convolution can be approximated to a normal distribution.

To derived the properties of the approximated normal distribution, we use the real mean and the variance of the convoluted function as follows.
\begin{align}
\begin{split}
{\rm E} \left( X_{c,1} \right) &= 
\frac{
\int_{-\infty}^{\infty}
p^i
P_c^1(p^i) dp^i
}
{
\int_{-\infty}^{\infty}
P_c^1(p^i) dp^i
	}
\\
&= 
\frac{
\int_{-\infty}^{\infty}
p^i
P_c^1(p^i) dp^i
}
{
\int_{-\infty}^{\infty} F_{ij}(p^i) dp^i \,
\int_{-\infty}^{\infty} P_{2}(p^i) dp^i 
}\\
&=
\frac{1}{V_\mathcal{B}}
\int_{-\infty}^{\infty}
	P_2(p^j) 
\int_{-\infty}^{\infty}
p^i
F_{ij}(p^j- p^i) 
	dp^i dp^j 
\end{split}
\end{align}
The denominator becomes the volume of the Minkowski sum of the two objects, $V_\mathcal{B}$, because the integral of the convolution in Eq. (\ref{eq:conv}) becomes the product of the integrations of both functions.

By defining a new variable $\eta \triangleq p^j - p^i$, we can further simplify as follows.
\begin{align}
\begin{split}
=&
\frac{1}{V_\mathcal{B}} 
\int_{-\infty}^{\infty}
	P_2(p^j) 
\int_{-\infty}^{\infty}
\left(p^j - \eta\right)
F_{ij}(\eta)
	d\eta\, dp^j  \\
=&
\frac{1}{V_\mathcal{B}} 
\bigg(
\int_{-\infty}^{\infty}
p^j
P_2(p^j) 
dp^j \,
\int_{-\infty}^{\infty}
F_{ij}(\eta)
	d\eta  \\
&
- \int_{-\infty}^{\infty}
	P_2(p^j) 
	dp^j\,
\int_{-\infty}^{\infty}
\eta
F_{ij}(\eta)
	d\eta
	\bigg) \\
=&
\mu_2
- 
\frac{1}{V_\mathcal{B}} 
\int_{-\infty}^{\infty}
\eta
F_{ij}(\eta)
	d\eta
\end{split}
\end{align}
We can define that the center of the Minkowski sum $F_{ij}$ is located at the origin, then the second term becomes zero and the mean of the conditional variable is the same as the mean of the second object position.

The covariance of $X_{c,1}$ can be derived from the expected value of $X_{c,1}X_{c,1}^T$ as follows.
\begin{align}
\rm cov (X_{c,1}) =& \rm E (X_{c,1}X_{c,1}^T ) - E(X_{c,1})^2
\end{align}
With the same definition of $\eta$, the expected value of $X_{c,1} X_{c,1}^T$ can be derived as follows.
\begin{align}
\begin{split}
& \rm E (X_{c,1} X_{c,1}^T ) \\ 
=&
\frac{
\int_{-\infty}^{\infty}
p^i\,{p^i}^T
P_c^1(p^i) dp^i
}
{
\int_{-\infty}^{\infty}
P_c^1(p^i) dp^i
	}\\
=&
\frac{1}{V_\mathcal{B}}
\int_{-\infty}^{\infty}
	P_2(p^j) 
\int_{-\infty}^{\infty}
p^i {p^i}^T
F_{ij}(p^j- p^i) 
	dp^i dp^j \\ 
=&
\frac{1}{V_\mathcal{B}}
\int_{-\infty}^{\infty}
	P_2(p^j) 
\int_{-\infty}^{\infty}
(p^j - \eta) (p^j - \eta)^T
F_{ij}(\eta) 
	d\eta\, dp^j \\ 
=&
\frac{1}{V_\mathcal{B}}
\int_{-\infty}^{\infty}
p^j {p^j}^T
	P_2(p^j) 
	dp^j 
\int_{-\infty}^{\infty}
F_{ij}(\eta) 
	d\eta
	\\ 
&- 2\frac{1}{V_\mathcal{B}}
\int_{-\infty}^{\infty}
p^j
	P_2(p^j) 
	dp^j
\int_{-\infty}^{\infty}
\eta^T
F_{ij}(\eta) 
	d\eta
	\\ 
& + \frac{1}{V_\mathcal{B}}
\int_{-\infty}^{\infty}
	P_2(p^j) 
	dp^j
\int_{-\infty}^{\infty}
\eta \eta^T
F_{ij}(\eta) 
	d\eta\\
=&
\left(\Sigma_2 + \mu_2 \mu_2^T\right) - \frac{2}{V_\mathcal{B}} \mu_2 
\int_{-\infty}^{\infty}
\eta^T
F_{ij}(\eta) 
	d\eta \\
& + \frac{1}{V_\mathcal{B}}\int_{-\infty}^{\infty}
\eta \eta^T
F_{ij}(\eta) 
	d\eta\\
\end{split}
\end{align}
The second term becomes zero because the center of $F_{ij}$ is located at the origin as mentioned above.

Therefore, the covariance of the conditional variable, $X_{c,1}$ can be simplified as follows.
\begin{align}
\rm cov(X_{c,1}) &= 
\Sigma_2 
 + \frac{1}{V_\mathcal{B}}\int_{-\infty}^{\infty}
\eta \eta^T
F_{ij}(\eta) 
	d\eta
\end{align}
Then, we can define a new random variable, $\tilde{X}_c$, with a normal distribution and its statistic properties are the same to those of $X_c$ as follows.
\begin{align}
\tilde{X}_{c,1} \sim \mathcal{N} \left( \mu_2,\, \Sigma_2 + C_{\mathcal{B}} / V_{\mathcal{B}} \right)
\end{align}

\subsection{Implementation considerations}
\label{sec:simple}
To predict an imminent collision with time index, $k_c$, we would ideally need to compute the instantaneous collision probability of objects using non-Gaussian distributions according to Eq. (\ref{eq:pic2}). This computation would take a large amount of time to finally predict the collision-free object distributions of Eq. (\ref{eq:pof}). Sampling-based probability derivations have been proposed for computational efficiency \cite{lambert2008}, however these computations remain expensive for real-time usage. Therefore we compromise on the accuracy of the estimation using other simplifications.

First, we simplify the instantaneous collision probability of Eq. (\ref{eq:pic2}), which consists of an integration over the considered environment and the shapes of the objects on it. By assuming the distribution of the objects to be normal, the computation becomes efficient. The integration operation then becomes a single normal distribution, which allows the use of pervasive efficient numerical solutions for the integration of multivariate normal distributions. The detailed approximation and computation of $\mathbf{p}_{ic}$ are explained in Appendix \ref{sec:a_pc}. If the object distributions are assumed to be normal, the collision probability of the two objects can be simplified as follows.
\begin{align}
\mathbf{p}_{ic}^{ij} &= \int_{\mathcal{B}} \mathbf{P}_{conv} \left( x \right) dx
\end{align}
where $\int_{\mathcal{B}} \left(\cdot\right)dp$ denotes the integration along the Minkowski sum of the i-th object and j-th object, and $\mathbf{P}_{conv}$ is a normal distribution with the following probability.
\begin{align}
\mathbf{P}_{conv} &\sim \mathcal{N} \left( 
		\mu_i - \mu_j,\ 
		\Sigma_i + \Sigma_j
\right)
\end{align}
The approximate integration of multivariate normal distributions around elliptic shapes is studied in \cite{sheil1977}.
Therefore, the instantaneous collision probability can be computed effectively, such that the conditional probability on the last term of Eq. (\ref{eq:pac2}) can be simplified. 

Second, we revisit the conditional probability density function $\mathbf{P}_{of}$ of Eq. (\ref{eq:pof}), and approximate it as a normal distribution. Initially, it has a normal distribution because it comes from the Kalman filter of Eq. (\ref{eq:X}).
\begin{align}
\mathbf{P}_{of,0}^{ij} = \mathcal{N} \left( \overline{x}^i, \Sigma_{x,0}^i\right)
\end{align}
Because $\mathbf{P}_{of,0}^{ij}$ is a normal distribution, the probability at the next iteration, $\hat{\mathbf{P}}_{of,1}^{ij}$ can be estimated as follows.
\begin{align}
\hat{X}_{f,1} \sim \mathcal{N} \left( A_p X_0, A_p \Sigma_{X,0}^{i} A_p^T + \Sigma_w^i \right) \label{eq:hatxf}
\end{align}
If we assume that all the subsequent probabilities are also normal distributions as shown in Appendix \ref{sec:xf}, any $\hat{X}_{f,k}$ can be predicted from its previous estimate as follows.
\begin{align}
\hat{X}_{f,k}^{ij} \sim \mathcal{N} \left( A_p X_{k-1}^{ij}, A_p \Sigma_{X_f,k-1}^{ij} A_p^T + \Sigma_w^i \right) \label{eq:hatxf}
\end{align}

\bibliographystyle{plain}
\bibliography{paper}
\end{document}